\documentclass[journal]{IEEEtran}
\usepackage[noend]{algorithmic}
\usepackage{balance}
\usepackage{hyperref}
\usepackage{subcaption}
\usepackage{graphicx}
\usepackage{mathematix}
\usepackage{amsmath}
\usepackage{bm}
\usepackage{cleveref}
\usepackage{todonotes}
\usepackage{wrapfig}

\usepackage{mathtools}
\usepackage[thmmarks,amsmath]{ntheorem}

\usepackage{pgf}

\usepackage{cite}
\theoremnumbering{arabic}
\theoremstyle{plain}
\theoremsymbol{}
\theoremheaderfont{\normalfont\bfseries}
\theorembodyfont{\upshape}
\newtheorem{remark}{Remark}

\theoremstyle{plain}

\newcommand{\pkg}[1]{\textsc{#1}}

\author{Ajay Suresha Sathya$^{1,2}$, Herman Bruyninckx$^{1, 3}$, Wilm Decr\'e$^{1, 2}$ and Goele Pipeleers$^{1, 4}$  % <-this % stops a space
\thanks{*The authors gratefully acknowledge support from Flanders Make through the Flanders Make SBO project - MULTIROB and from the Research Foundation Flanders (FWO) through the project G0D1119N. Flanders Make is the Flemish strategic research centre for the manufacturing industry.}% <-this % stops a space
\thanks{$^{1}$The authors are with the Division of Robotics, Automation and
Mechatronics in the Department of Mechanical Engineering, KU Leuven, Leuven, Belgium,  $^{2}$ DMMS-M Lab, Flanders Make, Leuven, Belgium, $^{3}$ TU Eindhoven, Netherlands, $^{4}$ Materialise NV, Leuven, Belgium.   Contact:{\tt\small ajay.sathya@kuleuven.be}}%
}

\title{Efficient Constrained Dynamics Algorithms based on an Equivalent LQR Formulation using Gauss' Principle of Least Constraint}

\begin{document}

\maketitle
\thispagestyle{empty}
\pagestyle{empty}
\begin{abstract}

%Efficient computation of constrained rigid dynamics is important for simulation, optimal control and learning methods.
We derive a family of efficient constrained dynamics algorithms by formulating an equivalent linear quadratic regulator (LQR) problem using Gauss' principle of least constraint and solving it using dynamic programming. 
Our approach builds upon the pioneering (but largely unknown) $O(n + m^2d + m^3)$ solver by Popov and Vereshchagin (PV), where $n$, $m$ and $d$ are the number of joints, number of constraints and the kinematic tree depth respectively. We provide an expository derivation for the original PV solver and extend it to floating-base kinematic trees with constraints allowed on any link.
 We make new connections between the LQR's dual Hessian and the inverse operational space inertia matrix (OSIM), permitting efficient OSIM computation, which we further accelerate using matrix inversion lemma. 
By generalizing the elimination ordering and accounting for \pkg{MuJoCo}-type soft constraints, we derive two original $O(n + m)$ complexity solvers. Our numerical results indicate that significant simulation speed-up can be achieved for high dimensional robots like quadrupeds and humanoids using our algorithms as they scale better than the widely used $O(nd^2 + m^2d + d^2m)$ LTL algorithm of Featherstone.
 The derivation through the LQR-constrained dynamics connection can make our algorithm accessible to a wider audience and enable cross-fertilization of software and research results between the fields. %, such as transfer of new results in data-driven control of systems with uncertainty. %\textcolor{red}{We further explore an alternate elimination ordering of the variables in the DP method compared to the original PV solver, which eliminates the need to explicitly compute the OSI matrix and can reduce the computational complexity from cubic to linear in the number of constraints.}
\end{abstract}

\section{Introduction}

Rigid body mechanics is a long-studied field with fundamental contributions already made in the 18th and 19th centuries.
Since the 1970s, robotics research has focussed on developing computationally efficient dynamics algorithms \cite{featherstone2014rigid}.
Initial motivation for this research was to enable real-time dynamic simulation and computed torque control on the slow computers of the 1970s. 
Despite significant processor clock-time improvements since then, computing dynamics efficiently remains a relevant problem because it can positively impact modern robotics applications involving model predictive control (MPC)
and reinforcement learning. Faster computation enables MPC control designers to increase the prediction horizon which usually improves optimality and stability properties of the MPC controller \cite{rawlings2017model}. It can speed up contact-aware online trajectory optimization \cite{posa2014direct,neunert2017trajectory} and also shorten long training times in reinforcement learning from simulations. Unsurprisingly, implementing  efficient dynamics simulators remains an active research area \cite{coumans2021,carpentier2021proximal,todorov2014convex,Lee2018,plancher2022grid}. %\textcolor{red}{Add MPC and RL references here}.
%\wilm{maybe also say something about speeding up simulations, for example for the entertainment industry, or in engineering to simulate the behavior of systems.}

However, efficient dynamics algorithms are typically complex with ``a steep learning curve'' \cite{baraff1996linear} and are not discussed in introductory robotics textbooks \cite{murray2017mathematical, lynch2017modern}. Consequently, robotics researchers often use dynamics algorithms (especially constrained dynamics algorithms) implemented in simulators as a black-box and are therefore unable to adapt or debug the algorithms to suit their applications. By deriving efficient constrained dynamics algorithms (CDA) as the solution of an equivalent equality-constrained linear quadratic regulator (LQR) problem, we believe that this paper makes efficient CDAs accessible to researchers with an optimization and control background. This includes many roboticists that are MPC practitioners due to the rising popularity of differential dynamic programming (DDP) style \cite{tassa2014control} algorithms. The optimization-based perspective as well as the LQR connection opens up possibilities for transfer of software and recent research results between the fields, especially the recent data-driven methods for safe control of systems with uncertain dynamics \cite{mesbah2022fusion}. Our derivation is also self-contained and does not assume prior knowledge of LQR derivation.

%\textcolor{red}{ not based on optimization. PV theoretically interesting because of the optimization-based formulation. LQR-structure (PV solver derivation much simpler than the previous Rodrigues paper). Also practical, and accessible to wide audience.}

\subsection{Related work}
The first efficient recursive algorithms, with $O(n)$ complexity in the number of joints, for computing  the unconstrained forward dynamics  were independently discovered by Vereshchagin \cite{vereshchagin1974computer}
and Featherstone \cite{featherstone1983calculation}. However, Vereshchagin's solver ``was way ahead of its time and languished in obscurity for a decade'' \cite{featherstone2000robot}.
Featherstone's insight involved efficiently propagating the solution of the Newton-Euler equations through the links, while Vereshchagin's approach was based on optimizing the Gauss' principle of least constraint \cite{gauss1829neues}
(a fundamental optimization-based formulation of classical mechanics) using dynamic programming (DP) \cite{bellman1966dynamic}.
Vereshchagin's idea is analogous to the standard textbook approach for solving the discrete-time linear quadratic regulator (LQR) problem using DP \cite[Chapter 1]{rawlings2017model},
which we will use in the rest of this paper.
Similar connection to the LQR problem was independently made in \cite{rodriguez1987kalman} by noting similarities between the Kalman filter and $O(n)$ recursive dynamics algorithms and this connection was further developed within a \textit{spatial operator algebra} (SOA) framework \cite{rodriguez1991spatial}, \cite{rodriguez1992spatial}, making efficient $O(n)$ dynamics algorithms accessible to researchers familiar with filtering theory. However, the SOA derivation is fairly complex, is performed over several papers and assumes strong familiarity with filtering theory literature and notation from 1960s and 1970s. Moreover, the SOA derivation does not permit a straightforward extension to constrained dynamics. Unlike SOA, our LQR approach starts with the optimization problem arising from first principles, includes motion constraints and readers will find our derivation to be a significantly simpler and more direct connection to LQR than \cite{rodriguez1987kalman}. % and uses the formuala for inward force recursion and outward acceleration from existing dynamics literature.  %, yet , we do not assume prior knowledge of LQR and

\begin{figure}[h]
  \begin{center}
    \vspace{-25pt}
    \vspace{0.5cm}
    \centering
    \begin{subfigure}{.49\linewidth}
      \includegraphics[width=\textwidth]{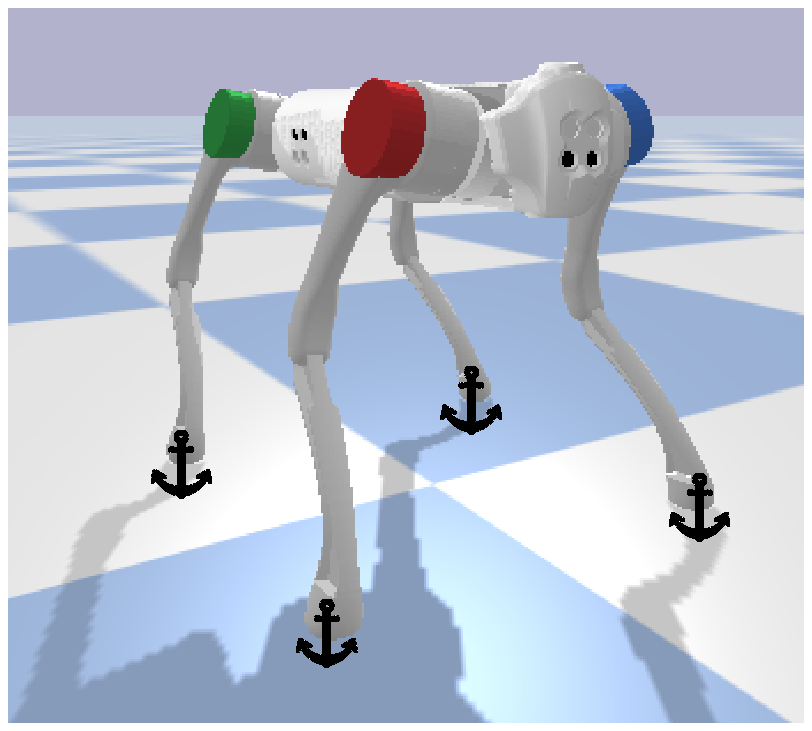}
      %\vspace{-17pt}
      \caption{Environment imposes constraints on a robot which must be accounted for in dynamics equations. }
      \label{fig:atlas_pybullet}
    \end{subfigure}\hspace{0pt}
    \vspace{-5pt}
    \begin{subfigure}{.49\linewidth}
      \includegraphics[width=\textwidth]{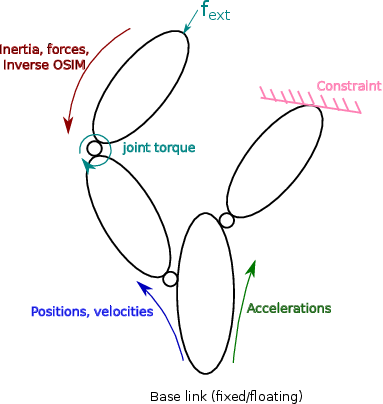}
      %\vspace{-17pt}
      \caption{Three sweep structure of the constraint dynamics solver paralleling forward simulation, backward DP recursion and rollout in LQR control.}
      \label{fig:kinematic_tree}
    \end{subfigure}
    \caption{Efficient computation of constrained dynamics by exploiting structure.}
    \label{fig:da_task}
  \end{center}
\end{figure}

%There has been renewed interest recently in computing rigid-body dynamics or/and their derivatives efficiently \cite{todorov2012mujoco}, \cite{Lee2018}, \cite{raisim}, \cite{carpentier2021proximal} and \cite{coumans2021}. Improvements in this regard can take contact-aware optimal control \cite{posa2014direct},  \cite{neunert2017trajectory} closer towards real-time model predictive control (MPC).

The simplicity arises from Gauss' principle allowing straightforward modeling of the motion constraints (like the non-penetration constraints for the feet of the Go1 robot in \cref{fig:atlas_pybullet}) by adding them as the constraints to the associated optimization problem. 
This ease of modeling allowed Popov and Vereshchagin (PV) to quickly extend their forward dynamics algorithm to an efficient $O(n)$ constrained dynamics algorithm \cite{popov1978manipuljacionnyje, vereshchagin1989modeling} for fixed-base kinematic chains with end-effector constraints. 
But this extension of the LQR connection to constrained dynamics remains largely unknown and unused by the robotics community despite its simplicity and efficiency. There have been a few robot control architectures using the PV solver \cite{ShakhimardanovAzamat2015CRMS, schneider2019exploiting}, including an implementation in \pkg{Orocos-KDL} \footnote{https://www.orocos.org/kdl.html} for kinematic chains, but its wider usage remains limited. \cite{ShakhimardanovAzamat2015CRMS} also derives the PV solver by introducing the concept of ``acceleration energy'' and extends it to trees by assembling acceleration energies. For readers unfamiliar with acceleration energy, their derivation is hard to follow and verify, while in this paper we provide an expository derivation purely using the mathematical perspective of dynamic programming on the LQR problem.

Other independent contributions that can be used to solve constrained dynamics includes the well-known operational-space formulation \cite{khatib1987unified}. However, \cite{khatib1987unified} does not propose an efficient algorithm for computing the operational-space inertia matrix (OSIM), which has a computational complexity of $O(n^3)$ when computed naively in joint-space. A major contribution to computing OSIM efficiently came in the form of an $O(n + m^2d + m^3)$ complexity recursive algorithm in \cite{kreutz1992recursive}, \cite{rodriguez1989spatial}, where $m$ is the number of constraints and $d$ is the tree depth. An efficient formula for computing off-diagonal blocks of the inverse OSIM using extended force propagators (EFP) was proposed in \cite{chang2001efficient}. However,  they do not exploit this EFP idea in their proposed algorithm and instead used a recursive approach similar to \cite{kreutz1992recursive} to obtain $O(n + mn + m^3)$ complexity \cite{chang2001efficient}.  The idea of EFP was fully exploited in the EFP algorithm (EFPA) \cite{wensing2012reduced} to obtain a reduced complexity of  $O(n + md + m^3)$. In \cite{featherstone2010exploiting}, Featherstone reported that exploiting the branching-induced sparsity in the joint-space inertia matrix (JSIM) and the kinematic Jacobian to compute the OSIM more efficiently than the existing recursive $O(n)$ algorithms despite having a worse $O(nd^2 + md^2 + dm^2)$ complexity (where $d$ is the depth of the tree) even for the Honda Asimo robot, a complex robot with $n = 40$. 
This result has led to a much wider usage of Featherstone's higher complexity method in rigid body dynamics  like \pkg{MuJoCo}, \pkg{Pinocchio} \cite{carpentier2021proximal}, Raisim and RBDL, to name a few, instead of the lower complexity EFPA algorithm \cite{wensing2012reduced}. Recent work \cite{carpentier2021proximal} derives Featherstone's OSIM algorithm \cite{featherstone2010exploiting} from the perspective of factorizing the contact KKT matrix and utilizes proximal-point iterations to solve for systems with redundant constraints.

%\textcolor{red}{This paper is limited to known acceleration constraints, which includes closed loops with ground.}

Independent efforts to extend the efficient ABA algorithm to internal kinematic closed loop constraints were realized in \cite{otter1987algorithm,bae1987recursive}. 
With the loop-closure constraint being a more general constraint model than the simpler desired acceleration-relative-to-ground constraint model considered in the PV solver, these more general algorithms include the PV solver computations as a subset of their computations. These algorithms can be straightforwardly adapted to kinematic trees with acceleration-relative-to-ground constraints to obtain an algorithm virtually identical to the PV solver.
The derivation in \cite{otter1987algorithm} relies heavily on the physical insight of the readers, while the derivation in \cite{bae1987recursive} is relatively more formal by algebraically solving the d'Alembert's equations. \cite{otter1987algorithm} further proposed a form of early constraint elimination that provides $O(m + n)$ complexity algorithm for certain kinematic mechanisms. Similar ideas were also used in an $O(n + m)$ complexity Lagrange multiplier-free algorithm \cite{anderson2003improvedorder} for certain kinematic mechanisms based on Kane's equations formulation of mechanics \cite{kane1985dynamics}. 
However, our PV solver derivation approach is different, and we will discuss in detail the comparison with these algorithms in \cref{sec:discussions_limitations}.  Moreover, we are not aware of any open-source implementation of \cite{otter1987algorithm,bae1987recursive} or its computational comparison with the popular Featherstone's sparsity exploiting algorithms.

Another line of research for accelerating dynamics computations includes the divide-and-conquer type of algorithms that aim to exploit parallel computing \cite{featherstone1999dividea,featherstone1999divideb,yamane2009comparative,bhalerao2012efficient} achieving an $O(\mathrm{log}(n))$ complexity provided that  $O(n)$ computational cores are used. These algorithms can be used to compute constrained dynamics by placing handles on the constrained bodies. The PV solver derived in \cref{sec:Tree_derivation} can also be similarly interpreted as an algorithm that computes the relative inertia of these handles. \cite{bhalerao2014distributed} presents a distributed algorithm specifically for computing the OSIM. Their comparison with this paper's algorithms are further discussed in \cref{sec:discussions_limitations}.

The efficient algorithms discussed so far have complex derivations, a third simple approach pioneered in \cite{baraff1996linear}, involves constructing the KKT matrix in `maximal' coordinates and solving it using a sparse linear solver. Despite having a favorable $O(n + md + m^3)$ complexity, Barraf's \cite{baraff1996linear} algorithm, does not exploit as much structure as possible (for example it computes joint constraint forces which are avoided in other methods) and requires joint constraint stabilization. It is generally not considered to be competitive with the recursive or sparse factorization methods mentioned above \cite{featherstone2014rigid}. 

%The PV solver (or \cite{otter1987algorithm} and \cite{bae1987recursive} with minor modifications) appears to exploit the constrained dynamics problem structure furthest. 
%But they have neither been compared nor computationally benchmarked against the currently popular sparse factorization methods or the recursive methods like KJR and EFPA, to the best of our knowledge, for the constrained dynamics problems considered in this paper. 
The PV solver derivation using our standard DP approach for LQR has the elegance and simplicity of Baraff's derivation, with a three-sweep structure that is analogous to forward simulation, backward DP recursion and rollout in LQR control as shown in \cref{fig:kinematic_tree}. We also found it be more efficient than state-of-the-art algorithms as we will show in the rest of this paper.

\subsection{Contributions}

%\textcolor{red}{Briefly introduce the contributions listed below}.

%We provide our derivation of the PV solver and propose several extensionsBut the PV solver first computes the dual function of the LQR problem, and the Hessian of this dual function is nothing but the inverse of the OSIM and this connection is unknown in the robotics and control literature. This also implies that it also includes within it the first known algorithm of $O(n)$ complexity for OSIM computation. \textcolor{red}{ This has not yet been computationally benchmarked with the other dynamics solvers to the best of our knowledge.} Additionally, the popular state-of-the-art articulated body algorithm (ABA) and efficient computation of the operational-space inertia matrix (OSIM) are special cases of the PV solver, and hence by-products of the derivation. which can allow transfer of ideas (for instance robust control algorithms) between the two fields. 

\subsubsection{Expository derivation of the original PV solver and extensions}
We provide an expository derivation of the original PV solver by adapting the textbook approach for solving the LQR problem \cite{rawlings2017model}, highlighting its connection to constrained dynamics more clearly than in existing literature. %We believe the derivation has value as the only English article \cite{vereshchagin1989modeling} on the PV solver presents the algorithm without derivation (we do not know their original derivation). %Moreover, due to the rising popularity of DDP and iLQR methods in robotics, our derivation is an accessible introduction to the PV solver to a large number of roboticists, not to mention researchers from optimization and control communities.
We then derive extensions to the original PV solver to support: 1) Floating-base robots 2) Constraints potentially on any link, 3) Kinematic trees and show its computational complexity to be $O(n + m^2d + m^3)$.%The derivation involves applying the dynamic programming on Gauss' principle of least constraint, two fundamental principles in optimization and analytical mechanics respectively. This leads to a significantly simpler derivation of constrained forward dynamics compared to the existing methods.
%The original solver \cite{vereshchagin1989modeling} considered only end-effector constrained kinematic chains.
% \wilm{did the original solver consider floating-bases, i.e. is this also an extension or not? I would write a little bit more explicit: we extend the solver in X ways: 1. support constrainsts on any link, 2. extend from chains to trees, ...}

\subsubsection{Connections to the OSIM}

We show that the dual Hessian, that is computed as an intermediate step of the PV solver, is equal to the inverse OSIM. This connection is new in literature, to the best of our knowledge, and provides an efficient $O(n + m^2d + m^2)$ algorithm, that is as yet unexploited to compute the inverse OSIM. 
This algorithm is structurally different from the currently known $O(n)$ family algorithms KRJ \cite{kreutz1992recursive} and EFPA \cite{chang2001efficient, wensing2012reduced}, by requiring only two sweeps over the kinematic tree instead of three and is found to be more efficient in practice for most robots of interest despite having a worse complexity than the $O(n + md + m^2)$ EFP algorithm. We further accelerate OSIM computation for floating-base robots with branching structure at the base. %\wilm{you mention that they have the same order of time complexity, but you do not say anything about the constant in front of O(n)... How do they compare here? For example, excluding more exotic algorithms, an SVD and an LU decomposition are both cubic... Especially for smaller matrices this constant is quite important...} 

%We demonstrate the linearity between accleration constraints and forces and the joint accelerations and segment accelerations. 

\subsubsection{$O(n + m)$ algorithms}

Building upon our expository PV solver treatment, we derive two efficient and new (to the best of our knowledge) constrained dynamics algorithms with only $O(n + m)$ computational complexity. 
The first algorithm solves the so-called ``soft Gauss principle'' used in the popular robot dynamics simulator \pkg{MuJoCo} \cite{todorov2012mujoco} \cite{todorov2014convex}, that relaxes the hard motion constraints with quadratic penalties. The second algorithm solves the original problem with hard motion constraints, by incorporating early elimination of Lagrange multipliers, thereby limiting their backward propagation which provides the improved computational complexity.

% \subsubsection{Soft Gauss' principle}

% \pkg{MuJoCo} \cite{todorov2014convex}, a popular robot dynamics simulator used widely in optimal control and reinforcement learning research, uses the so-called ``soft Gauss' principle'' where the motion constraints are relaxed with quadratic penalties, corresponding to the well known penalty method for approximately solving the constrained optimization problems \cite[Chapter 17]{nocedal2006numerical}. We adapt the PV solver to efficiently solve this problem and show that its computational complexity is similar to the articulated body algorithm.
%\wilm{maybe mention here something about (non-)exactness.}

% \subsubsection{Early resolution of multipliers} Using Bellman's principle of optimality, we generalize the elimination ordering of the primal and dual variables in the PV solver, thereby improving the computational performance.

% \subsubsection{Task-space hybrid dynamics} We propose a correction to the original PV algorithm to compute task-space hybrid dynamics (TSHD). %, which involves computing the joint torques (generalized forces) \wilm{the generalized forces between brackets reads a bit confusing: do you mean joint torques and/or external generalized forces (like: a person pushes the system with a certain force), or do you mean that the joint torques can also be joint forces (as in a prismatic joint)?} that must be exerted to achieve the desired motion in task space. %\textcolor{red}{This corresponds to the dynamically-consistent pseudoinverse. }

\subsubsection{Benchmarking} Despite the PV solver and Brandl et al.'s \cite{otter1987algorithm} contributions being over thirty-five years old, their computational performance is untested against the state-of-the-art algorithms, that are currently recognized to be fast in literature. We provide a comprehensive benchmarking of the PV solver against Featherstone's sparsity-exploiting algorithms \cite{featherstone2005efficient,featherstone2010exploiting} (currently used most widely in high-performance robot simulators including the \pkg{Pinocchio} and \pkg{MuJoCo} toolboxes), the lower-order EFPA \cite{chang2001efficient},\cite{wensing2012reduced} algorithm as well as our $O(n + m)$ extensions to the PV solver. These numerical results are new in literature to the best of our knowledge.

The source code of the solver is made available publicly~\footnote{\href{https://github.com/AjSat/spatial_V2}{https://github.com/AjSat/spatial\_V2}}.

\subsection{Organization}

We first discuss background material and preliminaries in \cref{sec:background} and derive the constrained dynamics solver for a kinematic chain with a fixed-base and motion constraints only on the end-effector in \cref{sec:derivation_chain}.  
We then discuss the physical interpretation of the terms of this relatively simple algorithm and also show the equality of the dual Hessian of the constrained LQR problem and the inverse OSIM in \cref{sec:physical_interpretation}.
Later, we generalize the derivation to the more complex case of floating-base robots with a kinematic tree structure and constraints on any link in \cref{sec:derivation_tree}. 
This separation of the PV solver derivation into two sections was made for clarity of exposition as it is easier to first follow the derivation for fixed-base kinematic chains before the generalization to trees. We then present an efficient extension of the PV solver to `soft' motion constraints in \cref{sec:soft_gauss}. We expand upon the dual Hessian-OSIM connection in \cref{sec:OSIM} and finish our derivations with a fast $O(n + m)$ algorithm for the original problem with hard motion constraints in \cref{sec:early_multiplier}.
Section~\ref{sec:experiments} presents algorithm benchmarking and discussions, and we make concluding remarks in \cref{sec:Conclusions}.

\section{Background} \label{sec:background}

\subsection{Notation and Convention} \label{sec: Notation}

Table~\ref{tab: notation} lists the notation used in this paper. Bold-faced lower case letters or symbols are vectors and upper case letters or symbols are matrices. 
$A^T$ is the transpose of a matrix $A$. 
$I_{n \times n}$ and $0_{n \times n}$ are the identity matrix and zero matrix of dimension $n \times n$ respectively. 
The $:=$ operator defines the left-side symbol with the right-side expression. 
The $\leftarrow$ operator assigns the right-side expression to a left-side variable in an algorithm.

We use the popular Featherstone's spatial algebra notation \cite{featherstone2014rigid} throughout the paper. For a robot's $i$th rigid body, $X_{i} \in SE(3)$,  $\mathbf{v}_i \in \mathbb{R}^6$ and $\mathbf{a}_i \in \mathbb{R}^6$ denote the spatial pose, velocity and acceleration respectively. $SE(3)$ is the special Euclidian group in 3 dimensions represented as a $6 \times 6$ spatial transformation matrix. 
$\mathbf{f}_i \in \mathbb{R}^6$ is the spatial force acting on the $i$-th body. 
%Refer \cite{featherstone2014rigid} for a detailed introduction to the spatial algebra notation.  
For notational simplicity of the upcoming derivations, all motion/force vectors $\mathbf{v}_i$, $\mathbf{a}_i$ and $\mathbf{f}_i$ are with respect to a common inertial frame. $\times$ and $\times^*$ are the spatial cross-product operators for motion vectors and force vectors respectively.

The whole robot's state is $(\mathbf{q_p}, \mathbf{\dot{q}})$, 
where $\mathbf{q_p} \in \mathcal{Q}$ is its pose in the configuration space $\mathcal{Q}$, $\mathbf{\dot{q}} \in \mathcal{T}_{q_p} {\mathcal{Q}} \simeq \mathbb{R}^n$ is its generalized velocity in $\mathcal{Q}$'s tangent space at $\mathbf{q}_p$ and $n$ is the robot's degrees of freedom (d.o.f).
Let $\boldsymbol{\tau} \in \mathcal{T}_{q_p}^{*} {\mathcal{Q}} \simeq \mathbb{R}^n$ be the generalized force acting on the robot in the dual tangent space of $\mathcal{Q}$ and $\mathbf{\ddot{q}} \in \mathbb{R}^n$ be $\mathbf{\dot{q}}$'s time derivative. 
This Lie algebraic notation allows a unified representation of floating-base robots and multi d.o.f joints where a singularity-free representation of position may require $n_p \geq n$. 
For a fixed-base manipulator with single d.o.f joints, $\mathbf{q_p} = \mathbf{q}$, $\mathbf{\dot{q}}$, $\mathbf{\ddot{q}}$ and $\boldsymbol{\tau}$ are simply the joint positions, velocities, accelerations and torques.

\begin{table}[t]
  \vspace{0.2cm}
  \caption{Notation \label{tab: notation}}
  \centering
  \begin{tabular}{l l}
    \hline
    Symbol                        & Definition                                                  \\
    \hline
    $^{\{j\}}X_i$                 & Spatial pose of $i$-th link in $j$-th link's frame.           \\
    $\mathbf{v}_{i}$              & 6D spatial velocity of the $i$-th link.                      \\
    $\mathbf{a}_i$                & 6D spatial acceleration of the $i$-th link.                  \\
    $\mathbf{f}_i$                & 6D spatial force acting on the $i$-th link.                  \\
    $\mathbf{q_p}$                & vector of robot joint positions.                            \\
    $\mathbf{\dot{q}}$            & vector of robot joint velocities.                           \\
    $\mathbf{\ddot{q}}$           & vector of robot joint accelerations.                        \\
    $\boldsymbol{\tau}$           & vector of robot joint torques.                              \\
    $n$                           & Degrees of freedom of the robot.                            \\
    $K_i$                         & Acceleration constraint matrix on $i$-th link.               \\
    $\mathbf{k}_i$                & Desired constraint accelerations.                           \\
    $J_i$                         & Geometric Jacobian of the $i$-th link.                       \\
    $\dot{J}_i$                   & Time derivative of $J_i$.                                   \\
    $J$                           & Joint-space constraint Jacobian.                            \\
    $\dot{J}$                     & Time derivative of $J$.                                     \\
    $\mathbf{k}$                  & Concatenation of all $\mathbf{k}_i$.                        \\
    $m$                           & Number of acceleration constraints on the robot.            \\
    $M$                           & Joint-space inertia matrix.                                 \\
    $\mathbf{c}$                  & Joint torques due to bias accelerations, forces and gravity. \\
    $\boldsymbol{\lambda}$        & Lagrange multipliers of constraints.                        \\
    $\Lambda$                     & Operational-space inertia matrix.                           \\
    $L$                           & Lower triangular matrix in LTL decomposition \cite{featherstone2005efficient}.  \\
    $Y$                           & Intermediate quantity in LTL-OSIM \cite{featherstone2005efficient}, see \cref{sec:featherstone_ltl}.  \\
    $\pi(i)$                      & Index of $i$-th link's parent link.                          \\
    $\gamma(i)$                   & Set of $i$-th link's children links' indices.                \\
    $S_i$                         & Motion subspace of the $i$-th joint.                           \\
    $T_i$                         & Force subspace of the $i$-th joint.                            \\
    $H_i$                         & $6 \times 6$ spatial inertia tensor of the $i$-th link.      \\
    $\mathbf{a}_{b,i}$            & $i$-th link's bias acceleration.                             \\
    $\mathcal{L}$                 & The Lagrangian of the LQR problem.                          \\
    %$\boldsymbol{\lambda}$ & Multipliers of cartesian-space constraints. \\
    $V_i$                         & Cost-to-go Lagrangian at $i$-th link.                        \\
    %$^{\{B\}}X_{\{C\}}^*$ & Transformation matrix for wrenches frame C to B \\
    $H_i^A$                       & Articulated body inertia of $i$-th link.                     \\
    $L_{i}^A$                     & Constraint's coupling due $i$-th and its descendant joints.  \\% Hessian of $V_i^*$ w.r.t to $\lambda_i^A$. \\
    $K_i^A$                       & Constraint force propagated to the $i$-th link.              \\
    $\mathbf{f}_i^A$              & Resultant force on $i$-th link excluding constraint forces.  \\
    $\mathbf{l}_i^A$              & Desired constraint accelerations propagated to the $i$-th link.  \\
    $D_i$                         & Apparent articulated body inertia along the $i$-th joint.                       \\
    $P_i$                         & Backward force propagator through the $i$-th joint.          \\
    $\mathbf{f}^{\mathrm{ext}}_i$ & Resultant external wrench acting on the $i$-th link.         \\
    $b$                           & Index of the floating-base link.                            \\
    $\boldsymbol{\lambda}_i^A$    & Concatenated multipliers on $i$-th and its descendant links. \\
    $r$                           & Number of branches from the floating-base link.             \\
    \hline
  \end{tabular}
\end{table}

\subsection{Preliminaries} \label{sec:preliminaries}

We will now briefly summarize forward dynamics, inverse dynamics and constrained dynamics problems. 
Forward dynamics computes $\mathbf{\ddot{q}}$, that result from applying $\boldsymbol{\tau}$ on a given robot at state $(\mathbf{q_p}, \mathbf{\dot{q}})$, to simulate the robot state forward in time. 
Conversely, inverse dynamics computes the $\boldsymbol{\tau}$ required to obtain a desired $\mathbf{\ddot{q}}$ at state $(\mathbf{q_p}, \mathbf{\dot{q}})$. Constrained dynamics is the forward dynamics problem with motion constraints in addition to joint constraints and will be formalized in the next paragraph. Inverse dynamics is, in general, easier to compute than forward dynamics, which is in turn significantly easier to compute than constrained dynamics. 

%Similarly to Featherstone's hybrid dynamics definition \cite[Chapter 9]{featherstone2014rigid}, that takes accelerations of a subset of joints and the torques about the remaining joints as inputs and computes the accelerations and torques for the whole robot, 

%We define TSHD (not a standard name) as the problem that  takes ``feed-forward'' torques $\boldsymbol{\tau}$ and task-space acceleration constraints as inputs and computes the resulting joint accelerations and the additional $\boldsymbol{\tau}_c$ that must be applied to satisfy task-space constraints. TSHD is typically computed using inertia-weighted pseudoinverse and nullspace projection in whole-body controllers, where the feedforward $\boldsymbol{\tau}$ is chosen to perform posture-control in the nullspace of constraints. 

Let the acceleration constraint on the $i$-th link be
\begin{equation} \label{eq:acc_con_i}
  K_i(\mathbf{q_p})\mathbf{a}_i = \mathbf{k}_i(\mathbf{q_p}, \mathbf{\dot{q}}),
\end{equation}
with $K_i \in \mathbb{R}^{m_i \times 6}$, $\mathbf{k}_i \in \mathbb{R}^{m_i}$ and $m_i$ the constraint dimensionality. 
Without loss of generality, we scale the constraints such that each row of $K_i$ has unit norm.
Both holonomic and non-holonomic motion constraints can be converted to this form by differentiation \cite{murray2017mathematical}.
The acceleration constraints can be transformed to the generalized coordinates using 
\begin{equation} \label{eq:geom_Jac}
  \mathbf{a}_i = J_i(\mathbf{q_p})\mathbf{\ddot{q}} + \dot{J}_i(\mathbf{q_p}, \mathbf{\dot{q}})\mathbf{\dot{q}},
\end{equation}
where $J_i(\mathbf{q_p}) \in \mathbb{R}^{6 \times n}$ is $i$th link's geometric Jacobian and $\dot{J}_i(\mathbf{q_p}, \mathbf{\dot{q}}) \in \mathbb{R}^{6 \times n}$ is its total time derivative. Substituting \cref{eq:geom_Jac} in \cref{eq:acc_con_i} and stacking all the links' constraints gives
\begin{equation} \label{eq:motion_constraint}
  J(\mathbf{q_p}) \mathbf{\ddot{q}} + \dot{J}(\mathbf{q_p}, \mathbf{\dot{q}})\mathbf{\dot{q}} = \mathbf{k}(\mathbf{q_p}, \mathbf{\dot{q}}),
\end{equation}
where $J(\mathbf{q_p}) := \begin{bmatrix}  K_1(\mathbf{q_p})J_1(\mathbf{q_p}) \\ \vdots \\  K_i(\mathbf{q_p})J_i(\mathbf{q_p}) \\ \vdots \\  K_n(\mathbf{q_p})J_n(\mathbf{q_p}) \end{bmatrix} \in \mathbb{R}^{m \times n}$,  $\dot{J}(\mathbf{q_p}, \mathbf{\dot{q}}) := \begin{bmatrix}K_1\dot{J}_1(\mathbf{q_p}, \mathbf{\dot{q}}) \\ \vdots \\ K_i\dot{J}_i(\mathbf{q_p}, \mathbf{\dot{q}}) \\ \vdots \\ K_n\dot{J}_n(\mathbf{q_p}, \mathbf{\dot{q}}) \end{bmatrix} \in \mathbb{R}^{m \times n}$, $\mathbf{k}(\mathbf{q_p}, \mathbf{\dot{q}}) := \begin{bmatrix} \mathbf{k}_1(\mathbf{q_p}, \mathbf{\dot{q}}) \\ \vdots \\ \mathbf{k}_i(\mathbf{q_p}, \mathbf{\dot{q}}) \\  \vdots \\ \mathbf{k}_n(\mathbf{q_p}, \mathbf{\dot{q}}) \end{bmatrix} \in \mathbb{R}^m$.

The constrained dynamics problem involves simultaneously solving \cref{eq:motion_constraint} and the linear system
\begin{align} \label{eq:constrained_dynamics}
  M(\mathbf{q_p})\mathbf{\ddot{q}} + \mathbf{c}(\mathbf{q_p}, \mathbf{\dot{q}}) +  J(\mathbf{q_p})^T\boldsymbol{\lambda} = \boldsymbol{\tau},
\end{align}
for unknowns $\mathbf{\ddot{q}}$ and $\boldsymbol{\lambda}$, where, $M(\mathbf{q_p}) \in \mathbb{R}^{n \times n}$,  $\mathbf{c}(\mathbf{q_p}, \mathbf{\dot{q}}) \in \mathbb{R}^n$ and $\boldsymbol{\lambda} \in \mathbb{R}^m$ are the joint-space inertia matrix (JSIM), generalized force due to Coriolis, centrifugal and gravity effects and the Lagrange multipliers associated with the acceleration constraint respectively.  Solving for $\mathbf{\ddot{q}}$ in \cref{eq:constrained_dynamics} (which is always possible because $M(\mathbf{q_p})$ is positive definite) and substituting in \cref{eq:motion_constraint} gives the operational-space form of constrained dynamics\cite{khatib1987unified} (with term dependencies dropped for brevity from now on when  it is clear from the context)
\begin{align} \label{eq: OSIM_Eom}
  \Lambda^{-1}\boldsymbol{\lambda} = \dot{J}\mathbf{\dot{q}} - \mathbf{k} + JM^{-1}(\boldsymbol{\tau} - \mathbf{c}),
\end{align}
with $\Lambda(\mathbf{q_p})^{-1} := (J(\mathbf{q_p})(M(\mathbf{q_p}))^{-1}J(\mathbf{q_p})^T) \in \mathbb{R}^{m \times m}$ and $\Lambda(\mathbf{q_p})$ is the OSIM. The inverse OSIM $\Lambda(\mathbf{q_p})^{-1}$ captures the inertial coupling between constraints, where the $i$-th column of $\Lambda(\mathbf{q_p})^{-1}$ is the acceleration along all the constraint directions caused by $\lambda_i = 1$ ($i$-th constraint force with unit magnitude). 
\begin{remark}
  Since $M(\mathbf{q_p})$ is a positive definite matrix, if $J$ has full row-rank, $\Lambda^{-1}$ has full rank, is invertible and $\Lambda$ exists. Then, \cref{eq: OSIM_Eom} permits a unique solution for $\boldsymbol{\lambda}$.
\end{remark}
 %It is possibleotherwise it permits either no solution (when motion constraints are infeasible) or an infinite number of solutions (when the system is over-constrained and there exists a nullspace of torques that do not result in any acclerations in the joint-space). \textcolor{red}{Are these claims necessary and also verify them}.
\begin{remark}
  $J$ may not have full row-rank in over-constrained systems, when constraints conflict with each other or due to loss of  $J_i$'s rank at kinematic singular configurations and depending on the numerical values of $\mathbf{k}_i$s, there exists either no solution or an infinite number of solutions for $\boldsymbol{\lambda}$.
 \end{remark}
Typical strategies to address singular $\Lambda^{-1}$ include Tikhonov regularization, proximal-point iterations \cite{carpentier2021proximal}, Moore-Penrose pseudo-inverse using the singular value decomposition (SVD), relaxing the constraints with weighted quadratic penalties \cite{todorov2014convex} or employing prioritized conflict resolution \cite{escande2014hierarchical}. Since a discussion of these different strategies is not the focus here, we assume that $J$ has full row-rank in the rest of this paper.

\subsection{Featherstone's LTL algorithms} \label{sec:featherstone_ltl}

We now review Featherstone's sparsity-exploiting algorithms and introduce terms that will be benchmarked in \cref{sec:experiments}. The LTL algorithm \cite{featherstone2005efficient} is a Cholesky decomposition for the JSIM 
\begin{equation}
  L^TL = M,
\end{equation}
where $L \in \mathbb{R}^{n \times m}$ is a lower triangular matrix. In contrast to the traditional LLT Cholesky algorithm \cite{golub2013matrix}, the LTL method ensures no fill-in (preserves the sparsity pattern of $M$ in $L$) even without resorting to pivoting methods that choose an elimination ordering. The idea was extended in the LTL-OSIM algorithm \cite{featherstone2010exploiting} to compute the OSIM for kinematic trees, where the sparsity pattern of $J$ is also exploited
\begin{equation}
  Y = JL^{-1},
\end{equation}
where $Y \in \mathbb{R}^{m \times n}$ also has the same sparsity pattern as $J$ and 
\begin{equation}
  \Lambda^{-1} = YY^T.
\end{equation}

\subsection{Forward kinematics}

\label{subsec:intro_fk}

Let a kinematic tree have $n$ links indexed from $1$ to $n$. 
The world link (assumed to be a fixed inertial frame) is assigned  the $0$ index. 
The $i$-th joint connects the $i$-th link to its parent link $\pi(i)$. 
The world link is tree's root and does not have a parent link. 
For floating-base robots, such as quadrupeds, a chosen link $b$ (usually the torso) is connected to the world link through a free joint. 
$\gamma(i)$ is the set of $i$-th link's children. A link $j$ is a leaf link if $\gamma(j) = \varnothing$.

The spatial poses, velocities and accelerations of all links in the tree can be computed recursively in a forward sweep starting from the root (world link) using
\begin{align}
  X_{j}        & = (X_{\pi(j)}) ({^{\{\pi(j)\}}X_{j'}}) ({^{{\{j'\}}}X_{j}}), \label{eq:fk}                                        \\
  \mathbf{v}_j & = \mathbf{v}_{\pi(j)} + S_j\mathbf{\dot{q}}_j, \label{eq:vel_prop}                                                \\
  \mathbf{a}_j & = \mathbf{a}_{\pi(j)} + S_j \mathbf{\ddot{q}}_j + \mathbf{v}_j \times S_j \mathbf{\dot{q}}_j, \label{eq:acc_prop}
\end{align}
where $^{\{\pi(j)\}}X_{j'}$ is the $j$-th link's pose in its parent link's frame when the $j$-th joint is at its home pose (usually computed from the robot URDF file or the DH parameters) and $^{{\{j'\}}}X_{j}$ is the spatial transformation due to the $j$-th joint's displacement. ${S_j} \in \mathbb{R}^{6 \times n_j}$ is the $j$-th joint's motion subspace, where $n_j$ is the joint's d.o.f (usually $1$). $S_j \mathbf{\dot{q}}_j$ is the $j$-th joint's contribution to $\mathbf{v}_i$. Let $T_j \in \mathbb{R}^{6 \times n_j}$ be the $j$-th joint's force subspace, such that $T_j \boldsymbol{\tau}_j$ is the joint's contribution to $\mathbf{f}_j$. 

\begin{remark}
  The force subspace $T_j$ is the dual of the motion subspace $S_j$, hence $S_j^TT_j = \mathbf{1}_{n_j \times n_j}$ \cite[eq. 3.39]{featherstone2014rigid}.
\end{remark}

\subsection{Gauss' Principle} \label{sec: GP}

Gauss' principle of least constraint \cite{gauss1829neues} (GPLC) is an optimization-based formulation of classical mechanics, which is not as well known or widely used as the Lagrangian formulation. 
Refer to \cite{udwadia1996} for a detailed discussion on Gauss' principle, according to which, a constrained system under the influence of forces undergoes accelerations that are as close as possible (in a weighted least-squares sense) to the unconstrained motion of the system under the same non-constraint forces.
For a system of rigid bodies with spatial inertia tensor $H_i \in \mathbb{R}^{6 \times 6}$ of the $i$-th link, under the external forces $\mathbf{f}_i$, which includes the bias forces $\mathbf{v}_i \times^* H_i\mathbf{v}_i$, the resulting accelerations $\mathbf{a}_i$ are the minimizers of the following optimization problem \cite{bruyninckx2000gauss}.
\begin{subequations}\label{eq:gauss_principle}
  \begin{align}
    \underset{\mathbf{a}_1, \hdots, \mathbf{a}_n}{\mathrm{\textbf{minimize}}}\quad & \sum_{i = 1}^{n}\frac{1}{2} (\mathbf{a}_i - H_i^{-1}\mathbf{f}_i)^T H_i(\mathbf{a}_i - H_i^{-1}\mathbf{f}_i),  \label{eq:GP-obj} \\
    \mathrm{\textbf{subject to}} \quad            & \mathrm{motion \ constraints}.
  \end{align}
\end{subequations}

%Deriving efficient constrained robot dynamics algorithms is significantly simpler and more direct by starting from Gauss' principle and using dynamic programming (DP) than all other existing derivations of constrained dynamics (that we are aware of), as we will show in the rest of the paper.

%Gauss' principle is at the same level as Newton's law. It is not as popular as D'Alembert's formulation or Euler-Lagrange equations and is not used as often in robotics textbooks and research literature for deriving robot dynamics, with Mujoco being an exception. Gauss' principle also has favourable theoretical properties, making it easily applicable to non-holonomic constraints \textcolor{red}{Read Udwadia properly and cite the chapter}. 

\subsection{Dynamic Programming Principle} \label{sec: DP}

Dynamic programming (DP) \cite{bellman1966dynamic} is a general theoretical framework for optimizing a function through a series of nested optimizations over the decision variables in some order. DP's efficiency can crucially depend on the variable elimination order. Each DP step optimizes over a function to return a function, so its implementation is intractable, unless the intermediate functions can be efficiently parameterized. The discrete-time linear quadratic regulator (LQR) problem is one such exception, where all the intermediate functions have the quadratic form. Fortunately, for kinematic tree mechanisms, the Gauss' principle is algebraically identical to the discrete-time LQR problem with scenario trees and can be solved efficiently using DP. This robot dynamics-LQR connection forms the basis of the derivations in this paper. %of efficient constrained dynamics algorithms in this paper.
%For illustration, suppose $f(x, y, z)$ is the objective function, which DP optimizes over the variables in some order. Without loss of generality, suppose $x$ is first optimized over to compute the function $g(y, z) = \underset{x}{\mathrm{\textbf{min}}} f(x, y, z)$, which is then optimized over y to compute function $h(z) = \underset{y}{\mathrm{\textbf{min}}} g(y, z)$.
%We then optimize $h(z)$ over $z$ to obtain $z^*$ and backsubstitute it to obtain $y^*$ and then $x^*$. 

\section{Derivation of the constrained dynamics solver} 
\label{sec:Derivation}

\label{sec:derivation_chain}

In this section we derive the PV solver for fixed-base kinematic chains with end-effector motion constraints. We first formulate the optimization problem in \cref{subsec:chain_formulation}, then derive its solution using DP in \cref{subsec:dp_chain_solution}. 

\subsection{Problem formulation} 

\label{subsec:chain_formulation}

Consider a kinematic chain with the links indexed such that $\pi(i) = i - 1$, with $0$-th link being the world link. 
The GPLC optimization problem \cref{eq:gauss_principle} for this chain is 

\begin{subequations}\label{eq:chain_gp}
\begin{align}
\underset{\mathbf{a}_1, \hdots, \mathbf{a}_n, \ddot{q}}{\mathrm{\textbf{minimize}}}\quad & \sum_{i = 1}^{n}\frac{1}{2} (\mathbf{a}_i - H_i^{-1}\mathbf{f}_i)^T H_i(\mathbf{a}_i - H_i^{-1}\mathbf{f}_i),  \label{eq:chain-obj}\\
   \mathrm{\textbf{subject to}} \quad & \mathbf{a}_{i} = \mathbf{a}_{i-1} + S_i\mathbf{\ddot{q}}_i + \mathbf{a}_{b,i}, \ i=1,2,...,n, \label{eq:chain_recurrence}\\
   & K_n\mathbf{a}_n = \mathbf{k}_n, \quad \mathbf{a}_0 = -\mathbf{a}_\mathrm{grav}, \label{eq:chain-con}
\end{align}
\end{subequations}
where \cref{eq:chain_recurrence} implicitly encodes joint motion constraints using \cref{eq:acc_prop}, $\mathbf{a}_{b,i} := \mathbf{v}_i \times S_i\mathbf{\dot{q}}_i$ is the bias acceleration, \cref{eq:chain-con} encodes the end-effector constraint (a common pattern e.g. when the end-effector is wiping a table) and the fixed-base constraint, and $\mathbf{a}_\mathrm{grav}$ is the acceleration-due-to-gravity vector. The reason for setting $\mathbf{a}_0$ to $-\mathbf{a}_\mathrm{grav}$  will be explained in \cref{subsec:gravity}. The parameters in the problem such as $H_i$, $\mathbf{f}_i$, $\mathbf{a}_{b,i}$ and $S_i$ are computed using the inputs to the problem, namely $\mathbf{q_p}$, $\mathbf{\dot{q}}$, $\boldsymbol{\tau}$ and the robot model.

The problem in \cref{eq:chain_gp} is algebraically identical to the discrete-time LQR problem: the forward propagation of link acceleration along the kinematic chain (see \cref{eq:chain_recurrence}) is analogous to the LQR's forward state propagation in time, with $\mathbf{a}_i$  and $\mathbf{\ddot{q}}_i$ corresponding to the LQR's states and controls respectively. 
\begin{remark}
Either $\mathbf{a_i}$s or $\mathbf{\ddot{q}}$ can be considered the \textit{free} variables in \cref{eq:chain_gp} as one can be computed from the other using \cref{eq:chain_recurrence} because $S_i$ always has full rank \cite{bae1987recursive}. 
 \end{remark}
\begin{remark}
The inertia tensor $H_i$ is positive definite for all links, therefore \cref{eq:chain_gp} is a strongly convex quadratic program (QP) with a unique solution, when feasible.
 \end{remark}
Conflicting constraints or unachievable desired accelerations at configuration $\mathbf{q}_p$ can make the QP infeasible. 

\subsection{Dynamic programming solution}

\label{subsec:dp_chain_solution}

We now solve the optimization problem in \cref{eq:chain_gp} using DP by following the textbook LQR derivation \cite[Chapter 1]{rawlings2017model}. The recurrence relation constraints in \cref{eq:chain_recurrence} and the $\mathbf{a}_0 = -\mathbf{a}_\mathrm{grav}$ constraint will be eliminated via substitution. However, unlike the textbook version, \cref{eq:chain_gp} has a hard `terminal' constraint (due to the end-effector constraint) which cannot be similarly eliminated via substitution. Therefore, we adapt the textbook derivation to instead solve for the primal-dual saddle point of QP's Lagrangian, which includes only the end-effector motion constraint as the joint and fixed-base constraints are eliminated through substitution
\begin{align} \label{eq:QP_chain_lagrangian}
&\mathcal{L}(\mathbf{\ddot{q}}, \boldsymbol{\lambda}) := \sum_{i = 1}^{n}\frac{1}{2} (\mathbf{a}_i - H_i^{-1}\mathbf{f}_i)^T H_i(\mathbf{a}_i - H_i^{-1}\mathbf{f}_i) + \\ & \qquad \qquad \qquad \boldsymbol{\lambda}^T(K_n\mathbf{a}_n - \mathbf{k}_n). \nonumber
\end{align}

We define ``cost-to-go Lagrangian" as the tail problem consisting of the Lagrangian terms corresponding to the $i$th link and its descendants 
\begin{align} \label{eq:QP_chain_costtogo}
&V_i(\mathbf{a}_{i-1}, \mathbf{\ddot{q}}_{i}, ... ,\mathbf{\ddot{q}}_n, \boldsymbol{\lambda}) := \nonumber \\ & \sum_{j = i}^{n}\frac{1}{2} (\mathbf{a}_j - H_j^{-1}\mathbf{f}_j)^T H_j(\mathbf{a}_j - H_j^{-1}\mathbf{f}_j) +  \boldsymbol{\lambda}^T(K_n\mathbf{a}_n - \mathbf{k}_n). \nonumber
\end{align}

Due to its additive structure, the cost-to-go Lagrangian follows the recurrence relation (after simplifying the quadratic objective and grouping the constant terms)
\begin{align} \label{eq:chain_costtogo_recurrence}
V_{i}(\mathbf{a}_{i-1}, \mathbf{\ddot{q}}_{i}, ... ,\mathbf{\ddot{q}}_n, \boldsymbol{\lambda}) =  \frac{1}{2} \mathbf{a}_{i}^T H_{i} \mathbf{a}_{i} -  & \mathbf{f}_{i}^T\mathbf{a}_{i} + \nonumber \\   \qquad V_{i+1}(\mathbf{a}_{i}, \mathbf{\ddot{q}}_{i+1}, ... ,&\mathbf{\ddot{q}}_n, \boldsymbol{\lambda)} + \mathrm{constant}. \nonumber
\end{align} 

When convenient, we will drop constant terms from now on for brevity. The Bellman's recurrence relation \cite{bellman1966dynamic} for the optimal cost-to-go Lagrangian is 
\begin{equation} \label{eq:chain_bellman}
V_{i}^*(\mathbf{a}_{i-1}, \boldsymbol{\lambda}) =  \underset{\mathbf{\ddot{q}}_i}{\mathrm{\textbf{min}}}\{\frac{1}{2} \mathbf{a}_{i}^T H_{i} \mathbf{a}_{i} -   \mathbf{f}_{i}^T\mathbf{a}_{i} +  V_{i+1}^*(\mathbf{a}_{i}, \boldsymbol{\lambda)} \}.
\end{equation} 

Optimizing the cost-to-go Lagrangian at the end-effector
\begin{equation} \label{eq:QP_chain_costtogo_n}
V_{n}(\mathbf{a}_{n-1}, \mathbf{\ddot{q}}_{n}, \boldsymbol{\lambda}) = \frac{1}{2} \mathbf{a}_{n}^T H_{n} \mathbf{a}_{n} - \mathbf{f}_{n}^T\mathbf{a}_{n} + \boldsymbol{\lambda}^T(K_n\mathbf{a}_n - \mathbf{k}_n),
\end{equation}
 over $\mathbf{\ddot{q}}_n$ gives $V^*_{n}(\mathbf{a}_{n-1}, \boldsymbol{\lambda})$. To do this, we first substitute $\mathbf{a}_n$ with the acceleration recursion equation in \cref{eq:chain_recurrence}
\begin{align}
V_{n}(&\mathbf{a}_{n-1}, \mathbf{\ddot{q}}_{n}, \boldsymbol{\lambda}) = \nonumber \\ & \frac{1}{2} (\mathbf{a}_{n-1} + S_n\boldsymbol{\ddot{q}}_n + \mathbf{a}_{b,n})^T H_{n} (\mathbf{a}_{n-1} + S_n\boldsymbol{\ddot{q}}_n + \mathbf{a}_{b,n}) - \nonumber \\ & \mathbf{f}_{n}^T(\mathbf{a}_{n-1} + S_n\boldsymbol{\ddot{q}}_n + \mathbf{a}_{b,n}) + \nonumber \\ & \boldsymbol{\lambda}^T(K_n(\mathbf{a}_{n-1} + S_n\boldsymbol{\ddot{q}}_n + \mathbf{a}_{b,n}) - \mathbf{k}_n). \label{eq:ctg_lagrangian_EE}
\end{align}
Then we collect the linear-quadratic terms in $\mathbf{\ddot{q}}_n$ and solve for the optimal $\mathbf{\ddot{q}}^*_n$, where the quadratic function's gradient is zero
\begin{align} 
  &\mathbf{\ddot{q}}_n^* = (S_n^T H_n S_n)^{-1}S_n^T\{\mathbf{f}_n - H_n(\mathbf{a}_{n-1} + \mathbf{a}_{b,n}) -  K_n^{T}\boldsymbol{\lambda}\}, \nonumber
\end{align}
substituting which back in \cref{eq:ctg_lagrangian_EE} provides $V^*_{n}(\mathbf{a}_{n-1}, \boldsymbol{\lambda})$, which remains a quadratic form in $\mathbf{a}_{n-1}$ and $\boldsymbol{\lambda}$. Therefore, let us hypothesize that $V_i^*(\mathbf{a}_{i-1}, \boldsymbol{\lambda})$  minimizes the following quadratic form 
\begin{subequations}
\begin{align} 
& V_i^*(\mathbf{a}_{i-1}, \boldsymbol{\lambda})  = \underset{\mathbf{\ddot{q}}_i}{\mathrm{\textbf{min}}} \{ \frac{1}{2}\mathbf{a}_i^TH_i^A\mathbf{a}_i - \frac{1}{2}\boldsymbol{\lambda}^TL_i^A\boldsymbol{\lambda} +  \label{eq:DP_ctg_beforesub}\\ 
& \quad \boldsymbol{\lambda}^TK_i^A\mathbf{a}_i -  \mathbf{f}^{AT}_i \mathbf{a}_i + \mathbf{l}_i^T \boldsymbol{\lambda}\} + \mathrm{constant} \nonumber \\
  &= \underset{\mathbf{\ddot{q}}_i}{\mathrm{\textbf{min}}}  \{\frac{1}{2}(\mathbf{a}_{i-1} + S_i\mathbf{\ddot{q}}_i + \mathbf{a}_{b,i})^TH_i^A(\mathbf{a}_{i-1} + S_i\mathbf{\ddot{q}}_i + \mathbf{a}_{b,i}) - \nonumber \\
& \qquad \frac{1}{2}\boldsymbol{\lambda}^TL_i^A\boldsymbol{\lambda} + \boldsymbol{\lambda}^TK_i^A(\mathbf{a}_{i-1} + S_i\mathbf{\ddot{q}}_i + \mathbf{a}_{b,i}) - \label{eq:DP_ctg_aftersub} \\ 
& \qquad \mathbf{f}^{AT}_i (\mathbf{a}_{i-1} + S_i\mathbf{\ddot{q}}_i + \mathbf{a}_{b,i}) + \mathbf{l}_i^T \boldsymbol{\lambda}\} + \mathrm{constant}. \nonumber 
\end{align}
\end{subequations}
where \cref{eq:DP_ctg_aftersub} is obtained by substituting \cref{eq:chain_recurrence} in \cref{eq:DP_ctg_beforesub}. Optimizing \cref{eq:DP_ctg_aftersub} over $\mathbf{\ddot{q}}_{i}$ by setting the objective function's gradient to zero gives
\begin{align} \label{eq:chain_optimal_qddot}
&\mathbf{\ddot{q}}_i^* = D_i^{-1}S_i^T\{\mathbf{f}_i^A - H_i^A(\mathbf{a}_{i-1} + \mathbf{a}_{b,i}) -  K_i^{AT}\boldsymbol{\lambda}\},
\end{align}
where $D_i^{-1} := (S_i^TH_i^AS_i)^{-1} \in \mathbb{R}^{n_i \times n_i}$ exists  because $S_i$ always has full column rank \cite{bae1987recursive}  and $H_i^A$ (which we will show to be the articulated body inertia matrix) is positive definite. Back-substituting $\mathbf{\ddot{q}}_i^*$ from \cref{eq:chain_optimal_qddot} in \cref{eq:DP_ctg_aftersub} gives $V_i^*(\mathbf{a}_{i-1}, \boldsymbol{\lambda})$, substituting which in the Bellman recurrence relation \cref{eq:chain_bellman} for $V_{i-1}^*(\mathbf{a}_{i-2}, \boldsymbol{\lambda})$ gives the following recursive formulae for the hypothesized quadratic form in \cref{eq:DP_ctg_beforesub},
\begin{subequations} \label{eq:chain_recursion}
\begin{align}
&H^A_{i-1} = H_{i-1} + P_iH_i^A,  \label{eq:ABA_recursion}\\
&\mathbf{f}^A_{i-1} = \mathbf{f}_{i-1} + P_i(\mathbf{f}^A_i - H_i^A\mathbf{a}_{b,i}), \label{eq:inward_force_prop}\\
&K^{AT}_{i-1} = P_{i}K^{AT}_{i}, \label{eq:constraint_force_prop} \\
&\mathbf{l}_{i-1} = \mathbf{l}_{i} + K_{i}^A\{\mathbf{a}_{b,i} + S_{i}D_{i}^{-1}S_{i}^T(\mathbf{f}_{i}^A - H_{i}^A\mathbf{a}_{b,i})\}, \label{eq:acc_setpoint_prop} \\
&L^A_{i-1} = L^A_{i} + K_i^AS_i(D_i)^{-1}S_i^TK_i^{AT}, \label{eq:OSM_Recursion}
\end{align}
\end{subequations}
where $P_i := (\mathbf{1}_{6\times 6} - H_i^AS_i(D_i)^{-1} S_i^T) \in \mathbb{R}^{6 \times 6}$ is the projection matrix that propagates forces and inertia backward through the $i$th joint. 

The end-effector cost-to-go Lagrangian in \cref{eq:QP_chain_costtogo_n} conforms to the hypothesized quadratic form in \cref{eq:DP_ctg_beforesub}, with $H_n^A = H_n$, $\mathbf{f}_n^A = \mathbf{f}_n$, $K_n^A = K_n$, $\mathbf{l}_n = -\mathbf{k}_n$ and $L_n^A = \mathbf{0}_{n \times n}$ being the starting point of the backward recursion using \cref{eq:chain_recursion}. With this, we can show inductively that the assumed quadratic form validly parameterizes the optimal cost-to-go-Lagrangian. 

Performing backward recursion until the root link yields $V_1^*(\mathbf{a}_0, \boldsymbol{\lambda})$'s expression, where the known value of $\mathbf{a}_0 = -\mathbf{a}_{\mathrm{grav}}$ is directly substituted, thereby eliminating all the primal variables of the Lagrangian to obtain the dual function

\begin{equation} \label{eq:dual_fun}
V_0^*(\boldsymbol{\lambda}) = -\frac{1}{2}\boldsymbol{\lambda}^TL_0^A\boldsymbol{\lambda} + \boldsymbol{\lambda}^T(\mathbf{l}_0 + K_0^A\mathbf{a}_0 ).
\end{equation}
Assuming that $L_0^A$ has full rank, the dual function has the unique maximizer
\begin{equation} \label{eq:opt_lambda_chain}
\boldsymbol{\lambda}^* = (L_0^{A})^{-1}(\mathbf{l}_0 + K_0^A\mathbf{a}_0).
\end{equation}
The numerical value of $\boldsymbol{\lambda}^*$ computed above enables rolling out the ``control policy" in a forward sweep to compute the optimal joint accelerations $\mathbf{\ddot{q}}_i^*$s using \cref{eq:chain_optimal_qddot} and \cref{eq:chain_recurrence}.

\subsubsection{Details on $\mathbf{f}_i$}

$\mathbf{f}_i$  is the resultant of all the non-constraint forces acting on the $i$th link, namely the force due to $i$th joint torque $\boldsymbol{\tau}_i$, the bias forces, the reaction force from $\boldsymbol{\tau}_{i+1}$ and all the other the external forces
\begin{equation} \label{eq:wrench on a link}
\mathbf{f}_i = T_{i}\boldsymbol{\tau}_i - \mathbf{v}_i \times^* H_i \mathbf{v}_i - T_{i+1}\boldsymbol{\tau}_{i+1} + \mathbf{f}_i^{\mathrm{ext}}.
\end{equation}

Note: the total reaction force on the $i$th link due to $\boldsymbol{\tau}_{i+1}$, must also include the backward propagation of the force acting on the $i+1$-th link due to $\boldsymbol{\tau}_{i+1}$,  $T_{i+1} \boldsymbol{\tau}_{i+1}$, using \cref{eq:inward_force_prop} in addition to the immediate reaction force $-T_{i+1}\boldsymbol{\tau}_{i+1}$,
\begin{align}
-T_{i+1}\boldsymbol{\tau}_{i+1} + &P_{i+1}(T_{i+1} \boldsymbol{\tau}_{i+1}) \\
& = -H_{i+1}^AS_{i+1}(D_{i+1})^{-1}S_{i+1}^TT_{i+1}\boldsymbol{\tau}_{i+1} \nonumber\\
& = -H_{i+1}^AS_{i+1}(D_{i+1})^{-1}\boldsymbol{\tau}_{i+1} \nonumber
\end{align}
which agrees with the known result on the backward reaction forces applied by joint actuators \cite[eq. 7.20]{featherstone2014rigid}.

\subsubsection{Including the effect of gravity}

\label{subsec:gravity}

The straightforward approach to account for gravity is to include the each link's weight in \cref{eq:wrench on a link}, but a more efficient and commonly used trick \cite{brandl1986very} is to add a gravity field by setting $\mathbf{a}_0 \leftarrow - \mathbf{a}_\mathrm{grav}$, where $\mathbf{a}_\mathrm{grav}$. Then $\mathbf{a}_i = - \mathbf{a}_\mathrm{grav}$ if the $i$th link is in equilibrium and $\mathbf{a}_i = 0$ if it is in free fall. This addition of gravitational acceleration to each link's acceleration must also be reflected the acceleration constraints through the update

$\mathbf{k}_n \gets \mathbf{k}_n - K_n\mathbf{a}_\mathrm{grav}$

\section{Physical interpretation}
\label{sec:physical_interpretation}

We will now provide the physical interpretation for the backward recursion in \cref{eq:chain_recursion}. This section is involved for readers not familiar with existing propagation-based constrained dynamics literature and may be skipped/skimmed during the first read. $P_i$ is the projection matrix, that propagates $\mathbf{f}_i$ through the $i$-th joint to the $i-1$-th link after removing the component that causes the $i$-th joint's motion. It is used in \cref{eq:inward_force_prop} to propagate the forces backwards in the chain. $P_i$ also propagates the inertia of the descendant links through the $i$-th joint in \cref{eq:ABA_recursion}, to compute the well known articulated body inertia $H_i^A$. Suppose that the $i$-th link was disconnected from its parent link but remained connected to its descendant links, $H_i^A$ would be this link's apparent inertia including the influence of all the descendant links. $D_i$ is the apparent inertia of the $i$-th link along the $i$-th joint, obtained by projecting $H_i^A$ onto the $i$-th  joint's motion subspace $S_i$.

In the absence of end-effector constraints, only \cref{eq:ABA_recursion} and \cref{eq:inward_force_prop} need to be computed during the backward recursion and these two formulae are identical to the inertia and force propagation equations in Featherstone's well known articulated body algorithm (ABA) \cite{featherstone1983calculation}, which remains the fastest algorithm to compute unconstrained forward dynamics \cite{featherstone2014rigid}. The PV solver reduces to ABA in the unconstrained setting and an unconstrained LQR-based derivation would essentially be an alternate derivation for the ABA algorithm. %Such a derivation (subset of the constrained version in the previous section) is mechanistic compared to the original ABA derivation \cite{featherstone1983calculation} that needed significant physical insight to efficiently and correctly propagate the solution of Newton-Euler equations through the chain.

%\textcolor{red}{This allows us to avoid having to compute the constraint forces in the directions where the joints are constrained.}

Each row of $K_n$ is the unit spatial force exerted by the end-effector due to the associated constraint, whose magnitude (the unknown Lagrange multipliers) must be solved for. These unit constraint forces are propagated backwards in the chain similarly to the non-constraint forces using the force propagator matrix $P_i$ in \cref{eq:constraint_force_prop}. Therefore, $-K_i^{A T} \boldsymbol{\lambda}$ is the force felt at the $i$-th link due to end-effector constraint forces.

Substituting the solution for joint accelerations from \cref{eq:chain_optimal_qddot} into the acceleration recurrence relation in \cref{eq:chain_recurrence} gives
\begin{equation} \label{eq:acc_proj}
   \mathbf{a}_i = P_i^T(\mathbf{a}_{i-1} + \mathbf{a}_{b,i}) + S_iD_i^{-1}S_i^T(\mathbf{f}_i^A - K_i^{AT}\boldsymbol{\lambda}),
\end{equation}
where $P_i^T$ is the projection operation that propagates $\mathbf{a}_{i-1}$ to child link $i$, after removing $\mathbf{a}_{i-1}$'s acceleration component along $S_i$. This reveals an interesting symmetric relationship between the forward acceleration propagator $P_i^T$ and the backward force propagator $P_i$ about the $i$-th joint, previously noted in \cite{lilly1989efficient}. Let us compose the force propagators to define the extended force propagator \cite{wensing2012reduced}
\begin{equation}
   P_i^n := P_iP_{i+1} ... P_{n}, \qquad \mathrm{and} \quad P_{n+1}^n := \mathbf{1}_{6 \times 6}
\end{equation}
that directly propagates end-effector forces to the $i-1$-th link. Due to the symmetric relationship, $P_i^{nT}$ propagates accelerations from the $i-1$-th link to the end-effector directly. Repeated substitution of \cref{eq:chain_optimal_qddot} for all joints in the acceleration recurrence relation \cref{eq:chain_recurrence} gives 
\begin{align}\label{eq:end_acc_sum_form}
   \mathbf{a}_n & = P_1^{nT}\mathbf{a}_{0} + \sum_{i = 1}^{n}   P_i^{nT} \mathbf{a}_{b,i} +                                               \\
                & \sum_{i = 1}^{n} \{P_{i+1}^{nT}  S_{i}D_{i}^{-1}S_{i}^T(\mathbf{f}_{i}^A - K_{i}^{AT}\boldsymbol{\lambda})\}.  \nonumber
\end{align}
From the constraint propagation equations in \cref{eq:constraint_force_prop}, one can easily verify that
\begin{equation} \label{eq:extended_con_force_prop}
   K_i^{A} = K_nP_{i+1}^{nT}.
\end{equation}

We remind readers that the end-effector acceleration constraint is $K_n\mathbf{a}_n + \mathbf{l}_n = 0$. Let us call $K_n\mathbf{a}_n$, \textit{constraint acceleration} (because it is the end-effector acceleration along the constrained direction) and  $-\mathbf{l}_n$ the desired constraint acceleration. Substituting $\mathbf{a}_n$ from \cref{eq:end_acc_sum_form} in the acceleration constraint equation and simplifying using \cref{eq:extended_con_force_prop} gives
\begin{subequations} \label{eq:end_acc_con_form}
   \begin{align}
      K_n\mathbf{a}_n + \mathbf{l}_n = K_0^{A}\mathbf{a}_{0} + \sum_{i = 1}^{n}   K_i^{A}P_i^T \mathbf{a}_{b,i} + \\
      \sum_{i = i}^{n} \{K_i^{A}S_iD_{i}^{-1}S_{i}^T(\mathbf{f}_{i}^A - K_i^{AT}\boldsymbol{\lambda})\} + \mathbf{l}_n = 0 . \nonumber
   \end{align}
\end{subequations}

$K_0^A \mathbf{a}_0$ is the constraint acceleration due to the known fixed-base acceleration. Collecting the terms not containing the unknown $\boldsymbol{\lambda}$ in the previous equation and comparing with backward recursion in \cref{eq:acc_setpoint_prop}, one can verify that
\begin{equation}
   \mathbf{l}_{i-1}^A = \sum_{k = i}^{n}  \{ K_k^{A}P_k^T \mathbf{a}_{b,k} +
   \{K_k^{A} S_{k} D_{k}^{-1}S_{k}^T(\mathbf{f}_{k}^A\}  \} + \mathbf{l}_n,
\end{equation}
recursively computes constraint acceleration caused by the bias accelerations, bias forces, joint torques and external forces up to the $i$-th joint and updates the desired constraint acceleration that must be supplied by the unknown constraint forces.  Comparing \cref{eq:OSM_Recursion} and \cref{eq:end_acc_con_form}, we see \cref{eq:OSM_Recursion} recursively computes the $\boldsymbol{\lambda}$-dependent terms in \cref{eq:end_acc_con_form} with
\begin{equation} \label{eq:L_i_accum}
   L_{i-1}^A = \sum_{k = i}^{n} K_k^A S_{k} D_{k}^{-1}S_{k}^T K_k^A
\end{equation}
where the $j$-th column of $L_{i-1}^A$ is the constraint accelerations caused by a unit magnitude $j$-th constraint force due to motions along the joints from the $n$-th joint back up to the $i$-th joint in the chain. $L_0^A$ represents the inertial coupling between constraints considering the whole tree's motion, providing intuition for why $L_0^A$ must be the inverse OSIM $\Lambda^{-1}$, which was previously defined in the joint-space in \cref{eq: OSIM_Eom}. %why . From this ex,  $L_0^A$  makes it intuitively easier to see why $L_0^A$ must be the inverse OSIM matrix. Finally the optimal magnitude of the constraint forces is obtained by solving the \cref{eq:opt_lambda_chain}.
\begin{equation} \label{eq:Lambda_jsim}
   \Lambda^{-1} = JM^{-1}J^T = K_n(J_nM^{-1}J_n^T)K_n^T,
\end{equation}
where $J_nM^{-1}J_n^T$ maps any force acting on the end-effector $\mathbf{f}_n$ to end-effector acceleration caused due to this force
\begin{equation} \label{eq:js_linmap}
   \mathbf{a}^f_n := (J_nM^{-1}J_n^T)\mathbf{f}_n.
\end{equation}

From \cref{eq:end_acc_sum_form}, we collect all the terms depending on $\mathbf{f_n}$ that cause end-effector acceleration (remember that $\mathbf{f}_i^A$ also depends on $\mathbf{f_n}$ because of inward force recursion ) to get
\begin{equation} \label{eq:os_linmap}
   \mathbf{a}^f_n =  \{\sum_{i = 1}^{n} P_{i+1}^{nT}  S_{i}D_{i}^{-1}S_{i}^TP_{i+1}^{n}\} \mathbf{f}_{n}.
\end{equation}

In \cref{eq:js_linmap} and \cref{eq:os_linmap} have linear mappings from $\mathbf{f}_{n}$ to $\mathbf{a}^f_n$, where $\mathbf{f}_{n}$ is free to take on any value in $\mathbb{R}^6$ and the linear mappings depend only on $\mathbf{q_p}$. Thus, it must be that $J_nM^{-1}J_n^T = \sum_{i = 1}^{n} P_{i+1}^{nT}  S_{i}D_{i}^{-1}S_{i}^TP_{i+1}^{n}$. Pre and post-multiplying this equality with $K_n$ and $K_n^T$,  we  get
\begin{align} \label{eq:inv_OSIM_eq}
   K_n(J_nM^{-1}J_n^T)K_n^T = \sum_{i = 1}^{n} K_nP_{i+1}^{nT}  S_{i}D_{i}^{-1}S_{i}^TP_{i+1}^{n}K_n^T,
\end{align}
where using \cref{eq:Lambda_jsim}, \cref{eq:extended_con_force_prop} and \cref{eq:L_i_accum}, we get $\Lambda^{-1} = L_0^A$. 
The physical interpretation presented here is essentially the argument used in \cite{otter1987algorithm} to derive their constrained dynamics solver for kinematic loops, which we refer readers to for more insight especially related to the effect of internal kinematic loops. Compared to \cite{otter1987algorithm}, our derivation is mathematical using the DP algorithm and does not require readers to possess physical insight. The physical interpretation provided here is only a post hoc explanation. However, the derivation in \cite{otter1987algorithm} does not assume prior optimization knowledge and may be more accessible to some readers, especially for those familiar with Featherstone's ABA algorithm derivation \cite{featherstone1983calculation} because \cite{otter1987algorithm} is a natural extension of \cite{featherstone1983calculation} that follows a similar variable elimination approach.  
%We showed this equality using the mapping end-effector force to end-effector acceleration, instead of the mapping from constraint forces to constraint accelerations, because constraint force weights $\boldsymbol{\lambda}$ may not exist when $\Lambda^{-1}$ is not invertible.

\section{Extension to trees with floating-base}
\label{sec:Tree_derivation}

\label{sec:derivation_tree}

We now extend the original PV solver, that only dealt with end-effector constrained fixed-base kinematic chains, to kinematic trees with possibly a floating-base  and possibly motion constraints on any link. We first modify the problem formulation to allow kinematic trees in \cref{subsec:tree_problem_formulation}, solve it using DP in \cref{subsec:tree_DP_solution} and finally present the algorithm and analyze the computational complexity in \cref{subsec:tree_algorithm}.

\subsection{Problem formulation}

\label{subsec:tree_problem_formulation}

The GLPC optimization problem for a given tree is
\begin{subequations}\label{eq:tree_gp}
\begin{align}
\underset{a, \ddot{q}}{\mathrm{\textbf{minimize}}}\quad & \sum_{i = 1}^{n}\frac{1}{2} (\mathbf{a}_i - H_i^{-1}\mathbf{f}_i)^T H_i(\mathbf{a}_i - H_i^{-1}\mathbf{f}_i),  \label{eq:tree_GP-obj}\\
   \mathrm{\textbf{subject to}} \quad & \mathbf{a}_i = \mathbf{a}_{\pi(i)} + S_i\mathbf{\ddot{q}}_i + \mathbf{a}_{b,i}, \quad i=1,2,...,n, \label{eq:tree_GP_recurrence}\\
   & K_i\mathbf{a}_i = \mathbf{k}_i, \quad i = 1,...,n, \label{eq:tree_GP_con}
\end{align}
\end{subequations}
where, $\pi(j)$ and $\gamma(j)$ are the parent link and the set of children for any given link $j$ respectively, as explained in \cref{subsec:intro_fk}. Compared to the problem in \cref{eq:chain_gp}, the recurrence relation in \cref{eq:tree_GP_recurrence} is indexed differently due to the tree structure, and any link's motion can be constrained in \cref{eq:tree_GP_con}. It is easily verifiable that the problem remains a strongly convex QP, but it is no more analogous to a simple discrete-time LQR problem. Instead, this problem shares its structure with scenario-trees from control of systems with dynamics uncertainty \cite{lucia2013multi}. However, the DP approach remains applicable and will provide a tree-structured Riccati recursion \cite{frison2017high}.

\subsection{Dynamic programming solution}

\label{subsec:tree_DP_solution}

Similarly to kinematic chains, we apply DP on the Lagrangian of the optimization problem in \cref{eq:tree_gp}
\begin{align} \label{eq:QP_lagrangian}
&\mathcal{L}( \mathbf{\ddot{q}}, \boldsymbol{\lambda}_1, ..., \boldsymbol{\lambda}_n) = \sum_{i = 1}^{n}\frac{1}{2}(\mathbf{a}_i^T H_i \mathbf{a}_i - \mathbf{f}_i^T\mathbf{a}_i) + \\ & \qquad \qquad \sum_{i = 1}^{n} \boldsymbol{\lambda}_i^T(K_n\mathbf{a}_i - \mathbf{k}_i). \nonumber
\end{align}

For notational simplicity in the upcoming derivation, let us define $\boldsymbol{\lambda}_i^A := [\boldsymbol{\lambda}_i^{T}, \boldsymbol{\lambda}^{AT}_{\gamma(i)_{1}}, \boldsymbol{\lambda}^{AT}_{\gamma(i)_{2}} ... \boldsymbol{\lambda}^{AT}_{\gamma(i)_{\mathcal{C}(i)}}]^T$ as the concatenation of the multipliers associated with constraints on the $i$-th link and its descendants, where $\mathcal{C}(i)$ is the cardinality of the set $\gamma(i)$. Analogously to the \cref{eq:chain_bellman}, the Bellman recurrence for the optimal cost-to-go Lagrangian for the kinematic tree is
\begin{align} \label{eq:tree_bellman}
V_{i}^*(\mathbf{a}_{\pi(i)}, \boldsymbol{\lambda}^A) =  & \underset{\ddot{q_i}}{\mathrm{\textbf{min}}} \{ \frac{1}{2} \mathbf{a}_i^T H_{i} \mathbf{a}_i -   \mathbf{f}_{i}^T\mathbf{a}_i + \boldsymbol{\lambda}_i^T(K_i\mathbf{a}_i - \mathbf{k}_i) + \nonumber \\  & \qquad \sum_{j \in \gamma(i)} V_{j}^*(\mathbf{a}_{i}, \boldsymbol{\lambda}_j^A) \} + \mathrm{constant}. \
\end{align}

Similarly to \cref{eq:DP_ctg_beforesub}, let us hypothesize that the optimal cost-to-go Lagrangian has the quadratic form
\begin{align} 
 V_i^*(\mathbf{a}_{\pi(i)}, \boldsymbol{\lambda}_i^A) & = \underset{\mathbf{\ddot{q}}_i}{\mathrm{\textbf{min}}} \{ \frac{1}{2}\mathbf{a}_i^TH_i^A\mathbf{a}_i - \frac{1}{2}\boldsymbol{\lambda}_i^{AT}L_i^A\boldsymbol{\lambda}_i^A + \label{eq:tree_DP_hypothesis}\\
& \boldsymbol{\lambda}_i^{AT}K_i^A\mathbf{a}_i - \mathbf{f}^{AT}_i \mathbf{a}_i + \mathbf{l}_i^T \boldsymbol{\lambda}_i^A \} + \mathrm{constant}. \nonumber 
\end{align}
Substituting $\mathbf{a_i}$ above using \cref{eq:tree_GP_recurrence} gives %the optimal cost-to-go Lagrangian as a function minimized over $\mathbf{\ddot{q}}_i$ analogously to \cref{eq:DP_ctg_aftersub}
\begin{align} 
 V_i^*(&\mathbf{a}_{\pi(i)}, \boldsymbol{\lambda}_i^A)  = \underset{\mathbf{\ddot{q}}_i}{\mathrm{\textbf{min}}} \{ \frac{1}{2}(\mathbf{a}_{\pi(i)} + S_i\mathbf{\ddot{q}}_i + \mathbf{a}_{b,i})^TH_i^A (\mathbf{a}_{\pi(i)} + \nonumber \\
 & S_i\mathbf{\ddot{q}}_i + \mathbf{a}_{b,i}) - \frac{1}{2}\boldsymbol{\lambda}_i^{AT}L_i^A\boldsymbol{\lambda}_i^A + \boldsymbol{\lambda}_i^{AT}K_i^A(\mathbf{a}_{\pi(i)} + \nonumber \\
 &S_i\mathbf{\ddot{q}}_i + \mathbf{a}_{b,i}) - \mathbf{f}^{AT}_i (\mathbf{a}_{\pi(i)} + S_i\mathbf{\ddot{q}}_i + \mathbf{a}_{b,i}) + \label{eq:tree_DP_hypothesis_after_sub} \\
&  \mathbf{l}_i^T \boldsymbol{\lambda}_i^A \} + \mathrm{constant}. \nonumber 
\end{align}

 Optimizing this function for optimal $\mathbf{\ddot{q}}_i$  gives
\begin{equation} \label{eq:tree_optimal_qddot}
\mathbf{\ddot{q}}_i^* = (D_i)^{-1}S_i^T\{\mathbf{f}_i^A - H_i^A(\mathbf{a}_{\pi(i)} + \mathbf{a}_{b,i}) -  K_i^{AT}\boldsymbol{\lambda}_i^{A}\},
\end{equation}
substituting which back into  \cref{eq:tree_DP_hypothesis_after_sub} gives $V_i^*(\mathbf{a}_{\pi(i)}, \boldsymbol{\lambda}_i^A)$. 

Substituting the expression $V_j^*(\mathbf{a}_{i}, \boldsymbol{\lambda}_j^A)$, thus computed for all $j \in \gamma(i)$ in Bellman recurrence relation \cref{eq:tree_bellman} confirms that the optimal cost-to-go function has the  quadratic form hypothesized in \cref{eq:tree_DP_hypothesis} for link $i$ if the hypothesis holds for all the children links $j \in \gamma(i)$. The quadratic form for the $i$-th link is given by the recursive equations
\begin{subequations} \label{eq:tree_backward_recursion}
\begin{align}
&H^A_{i} = H_{i} + \sum_{k \in \gamma(i)} P_kH_k^A,  \label{eq:tree_H_recursion}\\
&\mathbf{f}_i^A =  \mathbf{f}_i + \sum_{k \in \gamma(i)} P_k(\mathbf{f}_k^A - H_k^A\mathbf{a}_{b,k}), \label{eq:tree_f_recursion}\\
&K^A_{i} = \begin{bmatrix} K_i \\ $\vdots$ \\ K^A_{k}P_{k}^T \\ $\vdots$ \end{bmatrix}, \label{eq:tree_K_recursion} \\
&\mathbf{l}_i = \begin{bmatrix} -\mathbf{k}_i \\ \vdots \\ \mathbf{l}_{k} + K_{k}^A\{\mathbf{a}_{b,k} + S_{k}D_{k}^{-1}S_{k}^T(\mathbf{f}_k^A - H_{k}^A\mathbf{a}_{b,k})\} \\ \vdots \end{bmatrix},  \label{eq:tree_l_recursion} \\ 
&L^A_{i} = \begin{bmatrix} \mathbf{0}_{m_i \times m_i} & & & \\ & \ddots & & \\ & &L^A_{k} + K_k^AS_k(D_k)^{-1}S_k^TK_k^{AT} & \\ & & & \ddots \end{bmatrix}.  \label{eq:tree_L_recursion}
\end{align}
\end{subequations}

The cost-to-go Lagrangian at any leaf node $j$ is $V_j(\mathbf{a}_j, \boldsymbol{\lambda}^A_j) = \frac{1}{2}\mathbf{a}_j^TH_j^A\mathbf{a}_j - \mathbf{f}_j^{T}\mathbf{a}_j + \boldsymbol{\lambda}_j^T(K_j\mathbf{a}_j - \mathbf{k}_j)$. Thus, $H^A_j = H_j$, $L^A_j = \mathbf{0}_{m_j \times m_j}$, $K_j^A = K_j$, $\mathbf{f}_j^A = \mathbf{f}_j$, $\mathbf{l}_j = -\mathbf{k}_j$ for all $j$ that are leaf links. 
Therefore, it can be shown again inductively that the equations assumed in \cref{eq:tree_DP_hypothesis} correctly model the cost-to-go function. 

For a fixed-base robot, the backward recursion is performed until the base link $0$, and the known fixed-base acceleration is substituted to obtain the dual function, which is maximized to compute the optimal dual variables $\boldsymbol{\lambda}_0^{A*}$ (assuming that $L_0^{A}$ has full rank) analogously to \cref{eq:opt_lambda_chain}
\begin{equation} \label{eq:opt_lambda_tree}
\boldsymbol{\lambda}_0^{A*} = (L_0^{A})^{-1}(\mathbf{l}_0 + K_0^A\mathbf{a}_0).	
\end{equation}
%For a fixed-base robotic systems, the root node is usually chosen as the fixed-base. Therefore $a_0$ is known and can be directly substituted in \cref{eq:optimal_multiplier}. Often, a trick is employed to improve computational efficiency by choosing the fixed-base acceleration $a_0 = [0, \quad -g]$, where $g$ is the direction of the gravity vector. Then the gravitational force on each link need not be separately included in the $f_j$ terms. If this trick is employed, the computed acceleration at each segment is the sum of acceleration w.r.t to the base frame and the negative gravity vector. Hence, one must modify the desired acceleration term $k_n$ to $k_n := k_n + K_n^T[0, \quad -g]$ to maintain consistency.

For a floating-base robot, the backward sweep is conducted until the floating-base link $b$, from where the optimal base acceleration and the dual variables are the saddle point of the optimal cost-to-go Lagrangian at the floating-base
\begin{align}
\boldsymbol{\lambda_b}^{A*}, \mathbf{a}_b^* &= \mathrm{\textbf{arg}}\underset{\boldsymbol{\lambda}_b^A}{\mathrm{\textbf{max}}}\{ \underset{\mathbf{\mathbf{a}_b}}{\mathrm{\textbf{min}}} (\frac{1}{2}\mathbf{a}_b^TH_b^A\mathbf{a}_b - \frac{1}{2}\boldsymbol{\lambda}_b^{AT}L_b^A\boldsymbol{\lambda}_b^A + \\ 
&\boldsymbol{\lambda}_b^{AT}K_b^A\mathbf{a}_b -  \mathbf{f}^{AT}_b \mathbf{a}_b + \mathbf{l}_b^T \boldsymbol{\lambda}_b^A )\}. \nonumber 
\end{align}
 The stationary gradient condition of the first-order necessary KKT conditions provides the simultaneous linear equations,
\begin{align}
\mathbf{a}^*_b = &(H^A_b)^{-1}(\mathbf{f}^A_b - K^{AT}_b \boldsymbol{\lambda}^{A*}_b),\label{eq:tree_optimal_a}\\
\boldsymbol{\lambda}_b^{A*} = &(L_b^A)^{-1}(K_b^A\mathbf{a}^*_b + \mathbf{l}_b). \label{eq:tree_optimal_lambda}
\end{align}

We can substitute $\mathbf{a}^*_b$ from \cref{eq:tree_optimal_a} in \cref{eq:tree_optimal_lambda} to get
\begin{equation} \label{eq:floating_base_con_force}
\boldsymbol{\lambda}_b^{A*} = (L_b^A + K_b^A (H_b^A)^{-1} K_b^{AT})^{-1}(K_b^A(H_b^A)^{-1}\mathbf{f}_b^A + \mathbf{l}_b),
\end{equation}
and the optimal base acceleration is then recovered using \cref{eq:tree_optimal_a} and the inverse OSIM matrix is
\begin{equation} \label{eq:invOSIM_fb}
L_0^A = (L_b^A + K_b^A (H_b^A)^{-1} K_b^{AT}),
\end{equation}
which is no different from performing the usual backward recursion at the free-joint $b$ with, $S_b = I_{6 \times 6}$ as the free joint is allowed to move in all directions.

Alternately, if $L_b^A$ is invertible one can also substitute the expression for $\boldsymbol{\lambda}_b^{A*}$ from \cref{eq:tree_optimal_lambda} in to \cref{eq:tree_optimal_a} to get
\begin{align}\label{eq:floating_base_acc}
\mathbf{a}_b^* = (H_b^A + K_b^{AT}(L_b^A)^{-1}K_b^{A})^{-1}(\mathbf{f}_b^A-K_b^{AT}(L_b^A)^{-1}\mathbf{l}_b),
\end{align}
and optimal Lagrange multipliers can then be recovered using \cref{eq:tree_optimal_lambda}. The accelerations of the rest of the segments are then computed in the second forward sweep (rollout). The choice computing \cref{eq:floating_base_con_force} or \cref{eq:floating_base_acc} can significantly impact the computational efficiency of the algorithm depending on the branching structure and the number of constraints.

Suppose that kinematic tree branches at the floating-base, then $L_b^A$ has a block-diagonal structure because the $L_i^A$ terms from different branches occupy their respective diagonal block in \cref{eq:tree_L_recursion}. Factorizing or inverting $(L_b^A)^{-1}$ is easier due to this block-diagonal structure. Then computing \cref{eq:floating_base_acc} requires solving a small linear system of fixed size $6 \times 6$, which makes using \cref{eq:floating_base_acc} a superior choice in this case. On the other hand, computing \cref{eq:invOSIM_fb} performs a dense $m \times m$ update to $L_b^A$, which destroys the block-diagonal sparsity pattern and then requires solving a dense linear system of size $m \times m$.

\subsection{Algorithm}

Algorithm~\ref{alg:PV_tree} presents the PV solver for kinematic trees with floating-base. Let $\mathcal{S}$ be an ordered list of all the links in the kinematic tree, such that $i$ precedes $j$ in the list if $i$-th link is the $j$-th link's ancestor. Let $\mathcal{S}_r$ be the reversed list of $\mathcal{S}$. In 
\cref{alg:PV_tree}, we use \cref{eq:floating_base_acc} instead of \cref{eq:floating_base_con_force}.

\label{subsec:tree_algorithm}

\begin{algorithm}
\caption{PV solver for kinematic trees with floating-base}\label{alg:PV_tree}%
\begin{algorithmic}[1]
\REQUIRE 
\ $\mathbf{q^p}$,\
\ $\mathbf{\dot{q}}$,\
\ $\boldsymbol{\tau}$,\
\ $K_i$s,\
\ $\mathbf{k}_i$s,\
\ $X_{\{b\}}$,\
\ $\mathbf{v}_{\{b\}}$,\
\ robot model\

\textbf{First forward sweep}
\FOR {$i$ in $\mathcal{S}$}
\STATE $X_{\{i\}} = X_{\{\pi(i)\}} {^{\{\pi(i)\}}X_{{\{i'\}}}} {^{{\{i'\}}}X_{\{i\}}} $
\STATE $\mathbf{v}_i = \mathbf{v}_{\pi(i)} + S_i\mathbf{\dot{q}}_i$
\STATE $\mathbf{a}_{b,i} = \mathbf{v}_i \times S_i\mathbf{\dot{q}}_i$
\STATE $\mathbf{f}_i^A \gets \mathbf{f}_i^A + T_i\boldsymbol{\tau}_i - \mathbf{v}_i \times^* H_i \mathbf{v}_i + \mathbf{f}_i^\mathrm{ext}$; \ $K_i^A \gets K_i$; \ $\mathbf{l}_i \gets -\mathbf{k}_i; \ L_i^A \gets \mathbf{0}_{m_i \times m_i} \ H_i^A \gets H_i; $ \\
$\mathbf{f}_{\pi(i)}^A \gets \mathbf{f}_{\pi(i)}^A - T_i\boldsymbol{\tau}_i$
\ENDFOR{}
\textbf{Backward sweep}
\FOR {$i$ in $\mathcal{S}_r $}
\STATE $D_i = S_i^TH_i^AS_i; \ P_i = (\mathbf{1}_{6\times 6} - H_i^AS_i(D_i)^{-1} S_i^T)$
\STATE $\mathbf{f}^A_{\pi(i)} \gets \mathbf{f}_{\pi(i)}^A + 
P_i(\mathbf{f}_i^A - H_i^A\mathbf{a}_{b,i} )$
\STATE $H_{\pi(i)}^A \gets H_{\pi(i)}^A + P_iH_{i}^A$
\STATE $K_{\pi(i)}^A \gets \begin{bmatrix} K_{\pi(i)}^A \\  K_i^AP_i^T \end{bmatrix}$
\STATE $L^A_{\pi(i)} \gets \begin{bmatrix} L^A_{\pi(i)} & \\ & L^A_{i}  + K_i^AS_i(D_i)^{-1}S_i^TK_i^{AT} \end{bmatrix}$
\STATE $\mathbf{l}_{\pi(i)} \gets \begin{bmatrix} \mathbf{l}_{\pi(i)} \\ \mathbf{l}_{i} + K_{i}^A\{\mathbf{a}_{b,i} + S_{i}D_{i}^{-1}S_{i}^T(\mathbf{f}_{i}^A - H_{i}^A\mathbf{a}_{b,i})\} \end{bmatrix}$
\ENDFOR{}
\STATE $\mathbf{a}_b^* = (H_b^A + K_b^{AT}(L_b^A)^{-1}K_b^{A})^{-1}(\mathbf{f}_b^A-K_b^{AT}(L_b^A)^{-1}\mathbf{l}_b)$
\STATE $\boldsymbol{\lambda}^{A*}_b = (L_b^A)^{-1}(K_b^A\mathbf{a}^*_b + \mathbf{l}_b)$\\
\textbf{Second forward sweep (roll-out)}
\FOR{$i$ in $\mathcal{S}$}
\STATE $\mathbf{\ddot{q}}_i^* = (S_i^TH_i^AS_i)^{-1}S_i^T\{\mathbf{f}_i^A - H_i^A(\mathbf{a}_{\pi(i)}  + \mathbf{a}_{b,i}) -$ \\ $\qquad \qquad \qquad  K_i^{AT}\boldsymbol{\lambda}^{A*}_i\}$
\STATE $\mathbf{a}_{i} = \mathbf{a}_{\pi(i)} + S_i\mathbf{\ddot{q}}_i^* + \mathbf{a}_{b,i}$
\ENDFOR{}

\end{algorithmic}
\end{algorithm}

%\textcolor{red}{Algorithm explanation}

\subsubsection{Computational complexity}

We now analyze the worst-case computational complexity of \cref{alg:PV_tree}. The computations in lines 2, 3, 4, 5, 7, 8, 9, 17 each require fixed number of operations at every joint and requires $O(n)$ operations in total. The lines 10, 12, 16 require $O(m)$ operations per at most $d$ executions, where $d$ is the depth of the tree requiring $O(md)$ operations. Line 11 needs $O(m^2)$ operations per joint and $O(m^2d)$ operations in total. Factorizing $L_b^A$ in line 13  has the worst case complexity of $O(m^3)$. Aggregating these terms, the algorithm has requires $O(n + m^2d + m^3)$ operations in the worst case. 

\textit{Best case complexity:} The computational complexity is significantly better than the worst case complexity for favorable tree structures and constraints. Suppose that the branching occurs at the (floating) base link and there is one end-effector (a constrained link with at most 6 dimensional constraint) per branch. Quadrupeds and humanoid robots often have this structure. Let $r$ be the number of branches and $d$ be the length of the  longest branch. Line 11 is executed $d$ times for $r$ branches leading to $O(dr)$ operations. Similarly factorizing the block-diagonal matrix $L_b^A$ needs $O(r)$ operations for each block of size at most $6 \times 6$. As $m = O(r)$, the total complexity of the constrained dynamics for this tree is $O(n + md + m)$. %Suppose that for kinematic trees by employing sparse linear algebra routines that exploit the sparsity pattern are induced by branching \cite{featherstone2010exploiting}. For example, observe the diagonal block structure in \cref{eq:tree_L_recursion} where each of the blocks can be inverted in a decoupled manner.

The equality of $\Lambda^{-1}$ and $L_0$ established in \cref{sec:physical_interpretation} can be repeated for kinematic trees as well using identical arguments and hence will be skipped for the sake of brevity.

%\textcolor{red}{Anyways, it never makes sense to propagate $K_i^A$ or $L_i^A$ if the number of constraints is more than 6. Its better to just propagate an identity matrix instead of $K_i^A$ and later multiply $K_i^A$. Will explaining this complicate the paper?}

%\begin{align}
%&\mathbf{\ddot{q}}_j = (S_j^TH_i^AS_j)^{-1}(\boldsymbol{\tau}_j - S_j^T(H_j^A(a_{\pi(j)} + \ddot{X}_{b,j}) + \\ & \qquad \qquad \qquad \qquad \qquad \qquad  K_j^{AT}\mu_j + f_j^A) \nonumber \\
%&f^A_{j-1} = H_j^AS_jD_j^{-1}\boldsymbol{\tau}_j + P_j(H_j^A\ddot{X}_{b,j} + f^A_j + v_i \times^*H_iv_i)
%\end{align}

\section{Soft Gauss' principle}
\label{sec:soft_gauss}

We have considered only hard motion constraints so far, but it is also conceivable to relax these motion constraint through a penalty method and solve this easier problem, which is further always feasible even if the constraints are linearly dependent. This is precisely the approach taken in the \pkg{MuJoCo} toolbox \cite{todorov2012mujoco}, \cite{todorov2014convex}, a popular rigid body dynamics simulator using the so-called ``soft Gauss' principle'', where the hard motion constraints are relaxed through a quadratic penalty,
\begin{subequations}
\begin{align}
\underset{a, \ddot{q}}{\mathrm{\textbf{minimize}}}\quad & \sum_{i = 1}^{n}\frac{1}{2} \{(\mathbf{a}_i - H_i^{-1}\mathbf{f}_i)^T H_i(\mathbf{a}_i - H_i^{-1}\mathbf{f}_i) + \nonumber \\ & \qquad (K_i\mathbf{a}_i - \mathbf{k}_i)^T R_i^{-1}  (K_i\mathbf{a}_i - \mathbf{k}_i)\}, \label{eq:tree-obj-soft} \\
   \mathrm{\textbf{subject to}} \quad & \mathbf{a}_{i} = \mathbf{a}_{\pi(i)} + S_i\mathbf{\ddot{q}}_i + \mathbf{a}_{b,i}, \ i=1,2,...,n, \label{eq:tree_recurrence}
\end{align}
\end{subequations}
where $R_i \in \mathbb{R}^{m_i \times m_i}$ is a diagonal positive definite matrix.  After expanding the objective function in \cref{eq:tree-obj-soft}, collecting the quadratic and linear terms and ignoring the constant terms, we get an equivalent optimization problem,
\begin{subequations}
   \begin{align}
   \underset{a, \ddot{q}}{\mathrm{\textbf{minimize}}}\quad & \sum_{i = 1}^{n} \{ \frac{1}{2} \mathbf{a}_i^T(H_i + K_i^TR_i^{-1}K_i)\mathbf{a}_i \ - \nonumber \\ & \quad (\mathbf{f}_i + K_iR_i^{-1}\mathbf{k}_i)^T\mathbf{a}_i \} +  \mathrm{const}, \label{eq:tree-obj-soft_reformulated} \\
      \mathrm{\textbf{subject to}} \quad & \mathbf{a}_{i} = \mathbf{a}_{\pi(i)} + S_i\mathbf{\ddot{q}}_i + \mathbf{a}_{b,i}, \ i=1,2,...,n,
   \end{align}
   \end{subequations}
which is a special case of the kinematic tree optimization problem in \cref{eq:tree_gp}, but without motion constraints (apart from the joint constraints in \cref{eq:tree_recurrence} which will be eliminated through substitution) and with the modified $H_i$ and $f_i$ terms
\begin{equation}
H_i \gets H_i + K_i^TR_i^{-1}K_i; \quad \mathbf{f}_i \gets \mathbf{f}_i + K_iR_i^{-1}\mathbf{k}_i. \label{eq:inertia_forces_updates}
\end{equation}

As there are no motion constraints, the $L_i^A$, $\mathbf{l}_i$ and $K^A_i$ terms are not computed for the soft Gauss' problem, for which the algorithm~\ref{alg:PV_tree} reduces simply to ABA with the update in \cref{eq:inertia_forces_updates}.

\subsection{Computational complexity}

The ABA has $O(n)$ complexity while the inertia and forces updates in \cref{eq:inertia_forces_updates} require $O(m)$ operations. Therefore, the total computational complexity for solving the soft Gauss' principle is $O(m + n)$. %This reduced complexity does not depend on additional conditions like branching pattern or on whether the constraints are linearly independent as the soft-Gauss problem is always feasible.

The state-of-the-art simulator MuJoCo solves the problem in the joint-space resulting in a significantly higher computational complexity of $O(nd^2 + m^2d + d^2m)$. It uses the composite rigid body algorithm (CRBA) algorithm \cite[Method 3]{walker1982efficient} to compute the JSIM and factorizes it, which has worst-case complexity of $O(nd^2)$. It considers constraints by modifying the JSIM \cite[eq. 7]{todorov2014convex} analogously to our inertia update in \cref{eq:inertia_forces_updates} and solves this updated inertia matrix using the matrix inversion lemma accounting for the additional terms in the complexity. 

%\textcolor{red}{Add that this does not include the time-stepping, friction and other nasty constraints yet that Mujoco has. These features are outside the scope of the current paper.}

\section{$O(n)$ algorithm for OSIM}
\label{sec:OSIM}

The OSIM itself is an important expression in many rigid-body simulators in both the robotics and the computer graphics (where its inverse is known as the Delassus operator) communities. It also has applications in constrained inverse dynamics \cite{righetti2013optimal} and dynamically-consistent nullspace projection in prioritized torque control \cite{dietrich2015overview}. 
OSIM is particularly useful for resolving inequality constraints (also called unilateral constraints), because an inequality constraint becoming inactive can be easily handled by removing the corresponding row and column of the inverse OSIM and efficiently updating the factorization \cite{raisim}. 
Therefore, we isolate the OSIM computations in the PV solver and present a stand-alone algorithm. 
Further, we propose an at-best structure exploitation for floating-base robots that avoids factorizing the dense inverse OSIM, which all the existing approaches perform, to the best of our knowledge. 
Finally, we end the section with a qualitative comparison of the proposed algorithm with the existing $O(n)$ complexity OSIM solvers KJR \cite{kreutz1992recursive} and EFPA \cite{wensing2012reduced}.

\subsection{The PV-OSIM algorithm}

Algorithm~\ref{alg:PV_OSIM} lists the PV solver computations necessary for the OSIM.

\begin{algorithm}
  \caption{The PV-OSIM algorithm}\label{alg:PV_OSIM}%
  \begin{algorithmic}[1]
    \REQUIRE
    \ $\mathbf{q^p}$,\
    \ $K_i$s,\
    \ robot model\

    \textbf{First forward sweep}
    \FOR {$i$ in $\mathcal{S}$}
    \STATE $X_{\{i\}} = X_{\{\pi(i)\}} {^{\{\pi(i)\}}X_{{\{i'\}}}} {^{{\{i'\}}}X_{\{i\}}} $
    \STATE $K_i^A \gets K_i$; \ $ L_i^A \gets \mathbf{0}_{m_i \times m_i} \ H_i^A \gets H_i; $
    \ENDFOR{}
    \textbf{Backward sweep}
    \FOR {$i$ in $\mathcal{S}_r $}
    \STATE $D_i = S_i^TH_i^AS_i; \ P_i = (\mathbf{1}_{6\times 6} - H_i^AS_i(D_i)^{-1} S_i^T)$
    \STATE $H_{\pi(i)}^A \gets H_{\pi(i)}^A + P_iH_{i}^A$
    \STATE $K_{\pi(i)}^A \gets \begin{bmatrix} K_{\pi(i)}^A \\  K_i^AP_i^T \end{bmatrix}$
    \STATE $L^A_{\pi(i)} \gets \begin{bmatrix} L^A_{\pi(i)} & \\ & L^A_{i}  + K_i^AS_i(D_i)^{-1}S_i^TK_i^{AT} \end{bmatrix}$
    \ENDFOR{}
    \IF{floating-base?}
    \STATE $L_0^A = (L_b^A + K_b^A (H_b^A)^{-1} K_b^{AT})$
    \ENDIF{}
    \STATE $\Lambda = (L_0^A)^{-1}$\\
  \end{algorithmic}
\end{algorithm}

\subsection{The PV-OSIM-fast for floating-base robots}

\label{subsec:fast_osim}

For floating-base trees with branching at the base link, $L_b^A$ has block diagonal structure. This sparsity structure is lost in the update in line 10 in \cref{alg:PV_OSIM} (\cref{eq:invOSIM_fb}) by adding a dense matrix to $L_b^A$. The inverse OSIM (and the OSIM) is a dense matrix for floating-base robots because the constraints on different branches are coupled through the floating-base. All existing approaches, that we know of, compute this dense inverse OSIM and factorize it, which scales poorly in the presence of many constraints. We propose to avoid this by exploiting the structure of the update in \cref{eq:invOSIM_fb}.

%\textcolor{red}{Add in an algorithmic form}

%\textcolor{red}{floating-base destroys sparsity and makes OSIM computation very expensive and of the order $O(m^3)$. Here, we tackle this issue.}

The update to $L_b^A$ in \cref{eq:invOSIM_fb} is structurally a symmetric rank-6 update. If we assume that $L_b^A$ is invertible, which is a reasonable assumption for floating-base robots like humanoids and quadrupeds during operation, the matrix inversion lemma (MIL) \cite{sherman1950adjustment} can be used to factorize $L_0^A$ without having to explicitly construct this dense matrix. The MIL states
\begin{equation}
  (A + UCV)^{-1} = A^{-1} - A^{-1}U(C^{-1} + VA^{-1}U)^{-1}VA^{-1},
\end{equation}
applying which to solve \cref{eq:invOSIM_fb} yields
\begin{align}
  (L_0^A)^{-1} & = (L_b^A)^{-1} - (L_b^A)^{-1}K_b^A \{(H_b^A) +                                                  \\
               & K_b^{AT}(L_b^A)^{-1}K_b^{A}\}^{-1}K_b^{AT}(L_b^A)^{-1}, \nonumber                                    \\
               & = \Lambda_b - \L_K \{(H_b^A) + K_b^{AT}\L_K \}^{-1} \L_K^T \label{eq:structure_exploiting_osim}
\end{align}
where $\Lambda_b  := (L_b^A)^{-1}$ (easy to compute because of its block diagonal structure which is retained even after inversion) and $\L_K := \Lambda_b K_b^A$. Please note that the right-hand side (RHS) of the above equation is not evaluated to get the $(L_0^A)^{-1}$ matrix as that would destroy sparsity. Instead, the RHS is meant to be directly multiplied with vectors, similarly to how solving a linear system involves factorization and not matrix inversion.

\subsubsection{Computational complexity of PV-OSIM-fast}

The original PV-OSIM algorithm, computes and factorizes the dense $L_0^A$, which requires $O(m^3)$ operations. In contrast, the structure exploiting method computes $\Lambda_b$, which requires $O(\frac{m^3}{r^2})$ operations, and $L_K$, which requires $O(\frac{m^2}{r})$ operations, bringing the total complexity to $O(\frac{m^3}{r^2})$, where we have assumed for simplicity of analysis that the $m$ constraints are equally distributed among the $r$ branches. Thus, the proposed algorithm in this subsection can provide a significant speed-up for factorizing the inverse OSIM of floating-base robots with a favorable branching structure compared to the existing approaches that all solve dense linear systems.
%Multiplying it with a vector takes $O((\frac{m}{r})^2) + m)$ complexity compared to the original formula which has $O(m^2)$ complexity.  To use this form, the line \textcolor{red}{add this eqation in the algorithm} \cref{eq:invOSIM_fb} must be replaced with the following lines.

\subsubsection*{Limitation of PV-OSIM-fast}
Strictly speaking, PV-OSIM-fast is applicable in a subset of the cases where the regular PV-OSIM is applicable because of its assumption that $L_b^A$ is invertible. It is possible that $L_b^A$ is not invertible, but $L_0^A$ is invertible due to the addition of symmetric rank-6 matrix in \cref{eq:invOSIM_fb}. This situation may occur if there is a high dimensional constraint applied on a link close to the base link or if the robot reaches a kinematically singular configuration.

% \begin{subequations}
% \begin{align}
% \Lambda_b  &= (L_b^A)^{-1} \\
% \L_K &= \Lambda_b K_b^A\\
% (L_0^A)^{-1} &= \Lambda_b - \L_K \{(H_b^A)^{-1} + K_b^{AT}\L_K \}^{-1} \L_K^T \label{eq:structure_exploiting_osim}
% \end{align} 
% \end{subequations}

% Please note that explicitly computing $(L_0^A)^{-1}$ in \cref{eq:structure_exploiting_osim} which returns a dense OSIM matrix is unnecessary in almost all applications. It is numerically faster to directly multiply the right-hand side of \cref{eq:structure_exploiting_osim} for matrix-vector multiplications. Similar algorithms for limited rank updates for cholesky decomposition also exist, which can avoid operating on the dense inverse OSIM also exist in case one wants to factorize $L_0^A$ and not invert it, we refer interested readers to these \textcolor{red}{limited rank cholesky updates}. 

\subsection{Comparison with existing $O(n)$ OSIM algorithms}

\label{subsec:comparison_osim}

We now compare the PV-OSIM algorithm with the existing recursive $O(n)$ algorithms: the KJR algorithm \cite{kreutz1992recursive,rodriguez1992spatial}, whose optimized version was presented in \cite{featherstone2010exploiting}, and the \textit{extended force propagator} algorithm (EFPA) \cite{wensing2012reduced}. The three algorithms share the main idea of propagating the inverse inertia matrices, but differ significantly in the details. The primary structural difference of the PV-OSIM is that it computes the inverse OSIM in two sweeps while both KJR and EFPA require three sweeps. 

This difference arises because PV-OSIM computes inverse inertia due to the motion of the $i$th joint and its descendants directly in the \textit{constraint space} $L_i^A$  during the backward sweep, using the EFP to propagate constraint forces to a joint and the constraint accelerations back to the constrained link. 
However, both KJR and EFPA first compute the articulated body inertia in a backward sweep and then compute the spatial inverse inertia matrices of size $6 \times 6$ for all the necessary links in a forward sweep, which is avoided in the PV-OSIM.  
Propagating spatial inverse inertia matrices is a particularly expensive operation since they need to be transformed from one link's frame to another's (because dynamics algorithms are efficiently implemented in the link frame) in KJR and EFPA. This transformation is not required in PV-OSIM because the inverse inertia is directly computed in the constraint space.
Then KJR and EFPA compute the relative inverse inertia (essentially the matrix that maps forces on one link to the accelerations caused on another link) between every pair of links that are constrained. KJR performs computation inefficiently by propagating the relative spatial inverse inertia matrices through the path connecting two constrained links for every possible pair of constrained links.
EFPA computes the relative inverse inertia matrices more efficiently by directly transmitting the constraint forces and accelerations between constrained links through a common ancestor link using EFP. Finally, after all these inverse inertia matrices are computed, EFPA and KJR project them to the constraint space to get the inverse OSIM.

Thus, the PV-OSIM appears to exploit the structure of the problem better by using one less sweep to compute the inverse OSIM and its computational performance relative to existing OSIM algorithms will be benchmarked in \cref{subsec:benchmark_osim}. It must be noted that despite performing some extra computations, the EFPA algorithm has a lower order computational complexity of $O(n + md + m^2)$ compared to the $O(n + m^2d + m^2)$ complexity of the PV-OSIM for computing the inverse OSIM. Therefore, for kinematic trees of high depth and many constraints, we can expect the EFPA algorithm to be faster than the PV-OSIM, which we test in \cref{subsec:benchmark_osim}.

Also, note that the derivation of KJR or EFPA is complex and requires significant knowledge of and insight into efficient dynamics algorithms literature, while the PV-OSIM derivation is relatively simpler and self-contained as we are able to derive it from first principles (Gauss' principle) within this paper. Moreover, all the existing approaches compute and factorize the dense inverse OSIM matrix for floating-base robots, which the PV-OSIM-fast algorithm in \cref{subsec:fast_osim} avoids.

\section{Early multiplier resolution}
\label{sec:early_multiplier}

The original PV solver first eliminates the primal variables, recursively computes the inverse OSIM and factorizes it, which results in a worst case $O(n + m^2d + m^3)$ complexity. This can get particularly expensive when $m \sim O(n)$. However, if computing the OSIM is not required (for some other purpose during control or simulation), we can generalize the elimination ordering by aggressively eliminating dual variables earlier during the backward sweep to obtain an algorithm with an improved complexity of only $O(n + m)$. We now derive this algorithm by adapting our original PV solver derivation. This early elimination idea was already partly introduced in \cref{eq:floating_base_acc}, when we eliminated the dual variables just before eliminating $\mathbf{a}_b$ and will be further developed now. A form of early elimination is also proposed in \cite{otter1987algorithm}, where they eliminate the constraint forces of an internal kinematic loop as soon as all the link accelerations within that loop are eliminated. 

According to Bellman's principle of optimality \cite{bellman1966dynamic}, the solution to an optimization problem also optimizes its tail sub-problem. 
Hence, for the tail sub-problem at the $i$-th link
\begin{equation} \label{eq:Bellman_early_elim}
    \boldsymbol{\lambda}^{A*}_i = \underset{\boldsymbol{\lambda}^{A}}{\mathrm{\textbf{argmax}}} \quad V_{i}^*(\mathbf{a}_{\pi(i)}, \boldsymbol{\lambda}_i^A).
\end{equation}

%As a result, the dimensions of the matrix $L_0^A$ is $m \times m$, where the m is the total number of constraints. Since, both \cref{eq:optimal_multiplier} and \cref{eq:floating_base_acc} require factorizing $L_0$, the computational complexity associated with the original PV solver is $O(m^3)$. This can quickly become prohibitive in the presence of a large number of constraints. In this section, we an alternate order of elimination of the primal-dual variables to improve this complexity and see that we can even attain $O(m)$ for certain robot architectures.

The objective function above is of the form in \cref{eq:tree_DP_hypothesis} and is guaranteed to be bounded above and have a solution only when $L_i^A$ has full rank.  There is a rank-$n_i$ update to $L_i^A$ at every $i$-th joint during the backward recursion (see \cref{eq:tree_L_recursion})
\begin{equation} \label{eq:L_update}
    L^A_{i} \gets L^A_{i} + K_i^AS_i(D_i)^{-1}S_i^TK_i^{AT}
\end{equation}
Substituting the singular value decomposition (SVD) \cite{golub2013matrix} of  $L_i^A$ in \cref{eq:tree_DP_hypothesis} gives % and the dual feasibilty conditions after this update can be computed along the directions where the dual function lacks curvatureackward.
\begin{align} \label{eq:SVD_Bellman}
    \boldsymbol{\lambda}^{A*}_i = \underset{\boldsymbol{\lambda}^{A}}{\mathrm{\textbf{argmax}}} &\{ -\frac{1}{2}\boldsymbol{\lambda}_i^{AT} \begin{bmatrix}
        U_i^{1} & U_i^{2}   \end{bmatrix} \begin{bmatrix} \Sigma_i &   \\  & 0 \end{bmatrix} \begin{bmatrix} U_i^{1T} \\ U_i^{2T} \end{bmatrix} \boldsymbol{\lambda}_i^A + \nonumber\\ 
        & \mathbf{a}_i^TK_i^{AT}\begin{bmatrix}
            U_i^{1} & U_i^{2}   \end{bmatrix} \begin{bmatrix} U_i^{1T} \\ U_i^{2T} \end{bmatrix}\boldsymbol{\lambda}_i^{A} +  \\ & \mathbf{l}_i^T \begin{bmatrix}
            U_i^{1} & U_i^{2}   \end{bmatrix} 
        \begin{bmatrix} U_i^{1T} \\ U_i^{2T} \end{bmatrix}\boldsymbol{\lambda}_i^A \} + \mathrm{constant}, \nonumber 
\end{align}
where $\Sigma_i\in \mathbb{R}^{m_{ir} \times m_{ir}}$ is the diagonal matrix of the positive singular values, $U_i^{1} \in \mathbb{R}^{m_{if} \times m_{ir}}$ and $U_i^{2} \in \mathbb{R}^{(m_{if}) \times (m_{if}-m_{ir})}$ are the singular vectors corresponding to the positive and zero singular values of $L_i^A$, respectively, $m_{ir}$ and $m_{if}$ are the rank and the size of $L_i^A$, respectively. The left and right singular vectors are equal because $L_i^A$ is symmetric. Moreover, the singular vectors are orthonormal
\begin{equation}
\begin{bmatrix}
    U_i^{1} & U_i^{2}   \end{bmatrix} \begin{bmatrix} U_i^{1T} \\ U_i^{2T} \end{bmatrix} = I_{m_{if} \times m_{if}},
\end{equation}
which we use to project $\boldsymbol{\lambda}_i^A$, $K_i^A$ and $\mathbf{l}_i$ to two mutually orthogonal linear bases,  
\begin{align} \label{eq:SVD_projection}
    & \tilde{\boldsymbol{\lambda}}_i^A = U_i^{1T}\boldsymbol{\lambda}_i^A, \quad \hat{\boldsymbol{\lambda}}_i^A = U_i^{2T}\boldsymbol{\lambda}_i^A, \\
    & \tilde{K}_i^A = U_i^{1T}K_i^A, \quad \hat{K}_i^A = U_i^{2T}K_i^A,  \nonumber\\
    & \tilde{\mathbf{l}}_i = U_i^{1T}\mathbf{l}_i, \quad \hat{\mathbf{l}}_i = U_i^{2T}\mathbf{l}_i, \nonumber
\end{align}
where $\tilde{\boldsymbol{\lambda}}_i^A \in \mathbb{R}^{m_{ir}}$, $\tilde{K}_i^A \in \mathbb{R}^{m_{ir} \times 6}$, $\tilde{\mathbf{l}}_i \in \mathbb{R}^{m_{ir}}$ and $\hat{\boldsymbol{\lambda}}_i^A \in \mathbb{R}^{(m_{if}-m_{ir})}$, $\hat{K}_i^A \in \mathbb{R}^{(m_{if}-m_{ir}) \times 6}$, $\hat{\mathbf{l}}_i \in \mathbb{R}^{(m_{if}-m_{ir})}$ are the components of $\boldsymbol{\lambda}_i^A$, $K_i^A$ and $\mathbf{l}_i$ in the basis spanned by the singular vectors $U_i^{1}$ and $U_i^{2}$ respectively.
Using these quantities, the optimization problem in \cref{eq:SVD_Bellman} can be decoupled into a separate optimization problem and a dual feasibility condition along the columnspace and nullspace of $L_i^A$, respectively,
\begin{subequations}
    \begin{align}
        &\tilde{\boldsymbol{\lambda}}_i^A = \underset{\tilde{\boldsymbol{\lambda}}_i^A}{\mathrm{\textbf{argmax}}} \{-\frac{1}{2}\tilde{\boldsymbol{\lambda}}_i^{AT}\Sigma_i\tilde{\boldsymbol{\lambda}}_i^A + \mathbf{a}_i^T\tilde{K}_i^{AT}\tilde{\boldsymbol{\lambda}}_i^A + \tilde{\mathbf{l}}_i^T\tilde{\boldsymbol{\lambda}}_i^A \}, \label{eq:tilde_lambda} \\
        &\hat{K_i}^A\mathbf{a}_i + \hat{\mathbf{l}}_i = 0. \label{eq:hat_lambda}
     \end{align}
\end{subequations}

The solution to \cref{eq:tilde_lambda} is easily computed due to the diagonality of $\Sigma_i$,
\begin{equation}
\tilde{\boldsymbol{\lambda}}_i^{A*} = \Sigma_i^{-1}(\tilde{K}_i^A\mathbf{a}_i + \tilde{\mathbf{l}}_i). \label{eq:tilde_lambda_star}
\end{equation}

Substituting \cref{eq:tilde_lambda_star} back into the cost-to-go Lagrangian in \cref{eq:tree_DP_hypothesis} gives the following updates to its terms,
\begin{align}
    & H_i^A \gets H_i^A + \tilde{K}_i^{AT}\Sigma_i^{-1}\tilde{K}_i^A, \ \mathbf{f}_i^A \gets \mathbf{f}_i^A + \tilde{K}_i^{AT}\Sigma_i^{-1}\tilde{\mathbf{l}}_i, \nonumber \\
    & K_i^A \gets \hat{K}_i^A, \ \mathbf{l}_i \gets \hat{\mathbf{l}}_i, \ \boldsymbol{\lambda}_i^A \gets \hat{\boldsymbol{\lambda}}_i^A, \nonumber \\
    & L_i^A \gets 0_{(m_{if}-m_{ir}) \times (m_{if}-m_{ir})}. \label{eq:modified_cost_updates}
\end{align}

The backward recursion is performed using these modified terms in \cref{eq:tree_backward_recursion}. The early elimination is performed at each joint after $L_{(i)}^A$ is updated, which resets $L_{(i)}^A$ to zero matrix.
Early elimination reduces the number of propagated constraints at each $i$-th joint by $m_i$, which is the rank of $K_i^AS_i$ and usually equal to $n_i$, except in the case of redundant constraints or kinematic singularities. 
%Please note that $L_i^A$ is not propagated. 
If all the constraints are eliminated before reaching the root node, the backward sweep reduces to the ABA algorithm.

During the forward sweep, the optimal $\boldsymbol{\lambda}_i^{A*}$ is reconstructed using $\boldsymbol{\tilde{\lambda}}_i^{A*}$ from \cref{eq:tilde_lambda_star} and $\boldsymbol{\hat{\lambda}}_i^{A*}$ (available from the previous link) by transforming back to the original basis
\begin{equation}
\boldsymbol{\lambda}_i^{A*} = U_i^{1}\boldsymbol{\tilde{\lambda}}_i^{A*} + U_i^{2}\boldsymbol{\hat{\lambda}}_i^{A*}. \label{eq:lambda_star}
\end{equation}

For the common case of a single d.o.f joint, $L_i^A$ undergoes a rank-1 update in \cref{eq:L_update} and computing its SVD is computationally simple, with the singular vectors given by the following symmetric reflection matrix~\footnote{https://math.stackexchange.com/questions/704238/singular-value-decomposition-of-rank-1-matrix},
\begin{equation}
\begin{bmatrix} U_i^{1} & U_i^{2} \end{bmatrix} = I_{m_{if} \times m_{if}} - 2 \frac{\mathbf{w}_i\mathbf{w}_i^T}{\mathbf{w}_i^T\mathbf{w}_i}, \label{eq:SVD_rank1}
\end{equation}
and the positive singular value is 
\begin{equation}
\Sigma_i = \begin{bmatrix}
    \Vert \mathbf{ks}_i \Vert^2 / D_i 
\end{bmatrix}
\end{equation}
where 
\begin{equation} \label{eq:wi}
\mathbf{w}_i = \mathbf{ks}_i + \frac{ks_{i1}}{\vert ks_{i1} \vert} \Vert \mathbf{ks}_i \Vert \mathbf{e}_1, \ \mathbf{ks}_i = K_i^A S_i,
\end{equation}
where $ks_{i1}$ is the first element of $\mathbf{ks}_i$ and $\mathbf{e}_1$ is the first canonical basis vector. 

\begin{remark}
Since the rank-1 update SVD  can be computed using just $\mathbf{ks}_i$ and $D_i$, the $L_i^A$ matrix need not be explicitly updated. Furthermore, $U_i^1$ and $U_i^2$ matrices are not explicitly computed either because they are only needed for multiplying other matrices in \cref{eq:SVD_projection} and \cref{eq:lambda_star}, which is efficiently achieved by simply multiplying the right-hand-side of \cref{eq:SVD_rank1}. For example,
\begin{equation}
    \begin{bmatrix}
        \tilde{K}_i^A & \hat{K}_i^A 
    \end{bmatrix} = \{ I_{m_{if} \times m_{if}} - 2 \frac{\mathbf{w}_i\mathbf{w}_i^T}{\mathbf{w}_i^T\mathbf{w}_i} \} K_i^A. 
\end{equation}
\end{remark}

\begin{remark} The \cref{eq:wi} assumes $\mathbf{ks}_1 \neq 0$. If $\mathbf{ks}_1 = 0$, the rows of $\mathbf{ks}_i$ are permuted such that $ks_{i1} \neq 0$, similarly to the pivoting methods in matrix factorization algorithms \cite{golub2013matrix}.
 \end{remark}
 \begin{remark} If $\mathbf{ks}_i = 0_{m_{if} \times 1}$, the $i$th joint's acceleration is unaffected by the constraint forces $K_i^{AT}\boldsymbol{\lambda}_i^A$. In this case, the rank-1 update of $L_i^A$ in \cref{eq:L_update} would only add a zero matrix and is not performed. The terms $K_i^A$ and $\mathbf{l}_i$ are propagated to the parent link as in the original solver using \cref{eq:tree_K_recursion} and \cref{eq:tree_l_recursion} without size reduction.
 \end{remark}

%It is often possible (and likely) that $L_i^A$ reaches full rank before the backward recursion is completed, allowing early resolution of the Lagrange multipliers by solving \cref{eq:Bellman_early_elim},
% \begin{align} \label{eq:early_lambda}
%     \boldsymbol{\lambda}_i^{A*} = (L_i^{A})^{-1}(\mathbf{l}_i^A + K_{i}^A\mathbf{a}_i),
% \end{align}
% where $\boldsymbol{\lambda}_i^{A*}$ is an expression that is affine in $\mathbf{a}_i$. Substituting \cref{eq:early_lambda} back into the cost-to-go Lagrangian in \cref{eq:tree_DP_hypothesis} gives the following updates to the quadratic and linear coefficients of $\mathbf{a}_i$,
% \begin{subequations} \label{eq:modified_cost_updates}
%     \begin{align}
%         H_i^A \gets H_i^A + K_i^{AT} (L_i^{A})^{-1}K_i^A, \\
%         \mathbf{f}_i^A \gets \mathbf{f}_i^A + K_i^A (L_i^{A})^{-1} \mathbf{l}_i^A,
%     \end{align}
% \end{subequations}

\subsection*{Complexity analysis}

The PV-early solver's salient feature compared to the PV solver is that $L_i^A$ is not computed (hence $L_0^A$ is not factorized) and the matrices $K_i^A$ and $\mathbf{l}_i^A$ reduce in size during the backward sweep instead of growing with the accumulation of constraints. If the number of rows of $K_i^A$ and $\mathbf{l}_i^A$ is bounded by 6, the complexity of the PV-early solver is $O(n + m)$, as the number of operations at every joint is bounded by a constant. 

\begin{remark} If the  $K_i^A$ and $\mathbf{l}_i^A$ have more than 6 rows in the PV-early solver, it implies an over constrained system with more than 6 constraints on a link's acceleration. Then either the constraints are feasible with redundant constraints or infeasible, when one can remove the redundant constraints to obtain a constraint matrix $K_i^A$ with at most 6 rows or declare infeasibility early respectively.
\end{remark}

\section{Experiments and Discussion} \label{sec:experiments}

\label{sec:experiments}

We now benchmark and discuss the proposed algorithms. We 1) explain our implementation, 2)  benchmark the OSIM computation 3) benchmark the constrained dynamics algorithms themselves 4) empirically test the computational scaling of the different algorithms 5) discuss results and limitations of the proposed algorithms.

\subsection{Implementation}

We implemented the algorithms by extending Featherstone's highly readable MATLAB software toolbox SpatialV2 \cite{spatialV2}. For computing the OSIM, we implemented PV-OSIM and PV-OSIM-fast algorithms and to benchmark them we also implemented the KJR, EFPA and LTL \cite{featherstone2005efficient,featherstone2010exploiting} algorithms. For computing the constrained dynamics, we implemented PV, PV-early and the PV-soft algorithms and to benchmark them we also implemented the constrained dynamics algorithms using Featherstone's sparsity-exploiting LTL approach considering both the hard and the soft motion constraints. Robot specific C-code was generated for these algorithms using \pkg{CasADi}'s scalar expressions ({\tt{SX}}) \cite{andersson2019casadi} and its runtimes are used for the comparison. All the numerical experiments are performed on a single CPU core on a laptop with Intel i7-8850H CPU @ 2.60GHz processor running an Ubuntu 18.04 operating system. We disabled Intel {\tt Turbo Boost} during the benchmarking to reduce CPU frequency variability.

Implementing rigid body dynamics algorithms efficiently involves various nuances discovered by the robotics community over the years. 
For example, computing quantities in the local body frame instead of the inertial world frame can significantly reduce the number of operations needed \cite{brandl1986very}.
Thus our implementation also uses body frame though the derivation of the algorithms in this paper uses inertial frame for notational simplicity.
Also using the Denavit-Hartenberg (DH) structure for modelling the robot kinematics, whenever possible, makes the dynamics algorithms more efficient  \cite{mcmillan1995efficient}. However, this is not always possible, e.g. for kinematic trees, where a parent link can, in general, have DH structure with only one of the children joints. 
\cite{featherstone2005efficient,featherstone2010exploiting} carefully accounted for these nuances in their comparison of the LTL and ABA algorithms. 
Additionally, robot design can also significantly influence the operation count, e.g. some links  in the Kuka Iiwa have a 90-degree rotation between the parent joint's axis and the child joint's axis, resulting in a rotation matrix with only 3 non-zeros (either 1 or -1) requiring even fewer computations than DH nodes.
Therefore, an algorithm's operation count is robot-specific, and manually counting them for a given robot and constraint combination taking into account all the computational nuances would be tedious. Conveniently, \pkg{CasADi}'s {\tt{SX}} expression graph of an algorithm automatically provides the operation count allowing us to compare the best possible robot-specific operation count of the different algorithms, which we report later in this section.

 Our implementation further uses simple optimizations such as avoiding matrix-matrix operations whenever possible, and performing Cholesky factorization and solve instead of computing matrix inverses. The source code of the implementation~\footnote{\href{https://github.com/AjSat/spatial_V2}{https://github.com/AjSat/spatial\_V2}} and the simulation videos of the proposed algorithms~\footnote{\href{https://kuleuven-my.sharepoint.com/:f:/g/personal/ajay_sathya_kuleuven_be/EkuNpQF8BF5NhMiFajeiskIB0steWelFr_sxQkGa1P_Nrg?e=GS9SPg}{https://tinyurl.com/z78hkaah}} are made available. Baumgarte's stabilization was used in the simulations to stabilize the constraints over a long period of time \cite{baumgarte1972stabilization}, choosing a stabilization period of 0.1 seconds to avoid overly stiff dynamics as suggested in \cite[Section 8.3]{featherstone2014rigid}, which interested readers are referred to for further details.

 In our numerical experiments below, H and H$_3$ denotes a general 6D and 3D constraint on the `hand' link of a robot, with the corresponding $K_i$ being a random matrix of size $6 \times 6$ and $3 \times 6$ respectively.
 F and F$_3$ are defined similarly for the `foot' link. 
 For the Iiwa, the end-effector was considered the hand link. %, while for the Atlas and Talos the extreme link from the torso, being the gripper for Talos, was considered the hand link.

\subsection{Benchmarking the OSIM algorithms}

\label{subsec:benchmark_osim}

\begin{figure*}[t!]
\begin{subfigure}[h]{0.45\linewidth} 
    \centering
    \resizebox{\linewidth}{!}{\input{graphics/OSIM_atlas.pgf}} 
    \caption{Comparing the OSIM algorithms on the Atlas robot.}
    \label{fig:OSIM_atlas}
    \end{subfigure} 
    % \hspace{-1.05cm}
    \begin{subfigure}[h]{0.45\linewidth}
      \centering
      \resizebox{\linewidth}{!}{\input{graphics/OSIM_talos.pgf}}
      \caption{Comparing the OSIM algorithms on the Talos robot.}
      \label{fig:OSIM_talos}
      \end{subfigure}
      % \hspace{-1.2cm}
      \begin{subfigure}[h]{0.45\linewidth}
        \centering
        \resizebox{\linewidth}{!}{\input{graphics/OSIM_go1.pgf}}   
        \caption{Comparing the OSIM algorithms on the Unitree Go1 quadruped.}
        \label{fig:OSIM_go1}
        \end{subfigure} 
    \begin{subfigure}[h]{0.45\linewidth}
      \centering
      \resizebox{\linewidth}{!}{\input{graphics/OSIM_iiwa.pgf}}
      \caption{Comparing the OSIM algorithms on the KUKA Iiwa manipulator.}
      \label{fig:OSIM_kuka}
    % \end{center}
      \end{subfigure}
      \begin{subfigure}[h]{0.45\linewidth}
        \centering
        \includegraphics[width=\textwidth]{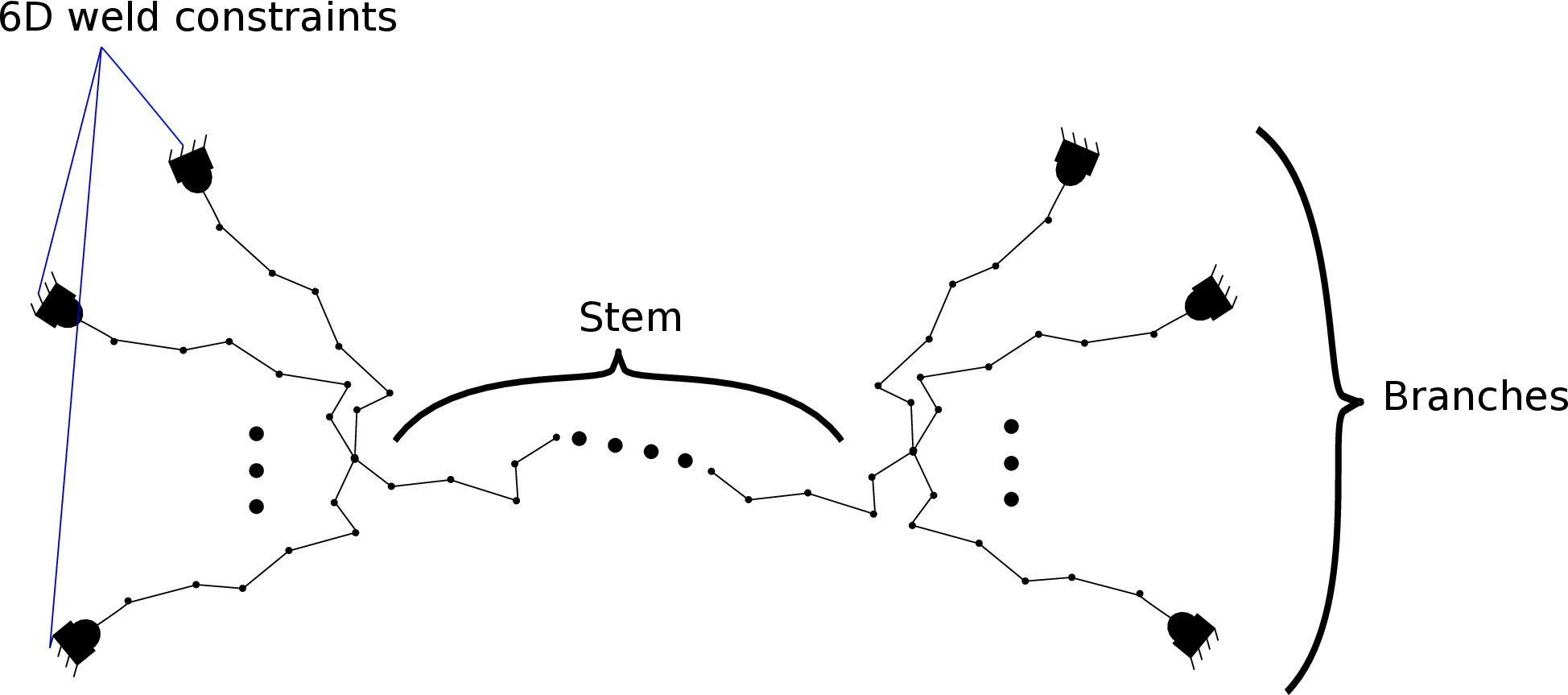}
        \caption{The long stalk on which the computational scaling is studied.}
        \label{fig:OSIM_stalk}
      % \end{center}
        \end{subfigure} \hspace{1.7cm}
      \begin{subfigure}[h]{0.45\linewidth}
        \centering
        \resizebox{\linewidth}{!}{\input{graphics/OSIM_long_tree.pgf}}
        \caption{Computational scaling of PV-OSIM vs EFPA. The key indicates the number of branches on either side of the long stalk for the mechanism in \cref{fig:OSIM_stalk}.}
        \label{fig:OSIM_long_tree}
      % \end{center}
        \end{subfigure}  
  \caption{Benchmarking the number of computation operations of the OSIM algorithms for various robots.}
  \label{fig:OSIM_benchmarks}
\end{figure*}

% \begin{figure}
%   \centering
%   \input{graphics/OSIM_atlas.pgf}
%   \caption{Example figure produced with this method.}
% \end{figure}

% \input{graphics/OSIM_atlas.pgf}

% add a table below listing the computational complexities of different OSIM algorithms

Figure~\ref{fig:OSIM_benchmarks} gives the operation count along with internal break-down for the proposed PV-OSIM and PV-OSIM-fast algorithms along with the existing SOTA $O(n + md + m^3)$ EFPA algorithm \cite{ wensing2012reduced} and the SOTA sparsity-exploiting $O(nd^2 + m^2d + dm^2)$ LTL-OSIM algorithm \cite{featherstone2010exploiting}. Similarly to \cite{wensing2012reduced}, we found KJR to be significantly slower than EFPA for all the considered robots and KJR would also scale worse due to its higher complexity, hence we omit the KJR results.  

We found PV-OSIM to be more efficient than the EFPA for all the considered robots. With the computation of articulated body inertia $I^A$, the task-space EFP $K^A$ and the Cholesky decomposition of inverse OSIM requiring the same number of computations for both algorithm, the difference arises in the inverse OSIM $\Lambda^{-1}$ computation. This is because EFPA requires an additional forward sweep, that propagates inverse inertia matrices forward with expensive similarity transformations, unlike the PV-OSIM as discussed in \cref{subsec:comparison_osim}.  

LTL-OSIM was the fastest algorithm for the KUKA Iiwa, which has only 7 d.o.f. However, for the 18 d.o.f Go1 robot the PV-OSIM was already slightly faster than the LTL-OSIM due to its lower computational complexity. For bigger robots like the Atlas (37 d.o.f) and Talos (50 d.o.f), LTL was the slowest of all the considered algorithms due to its higher computational complexity. A major difference between the LTL vs EFPA comparison in \cite{wensing2012reduced} (which found EFPA to be slower than LTL for the Honda Asimo robot) and ours is that we also include the cost of computing the constraint Jacobian computation $J$  in the LTL algorithm. We believe this to be a fairer comparison because the PV-OSIM and EFPA algorithms do not require $J$. 
$K^A$ propagates forces and accelerations from end-effectors to other links fulfilling a role similar to $J$ in LTL.
For fewer number of constraints, both PV-OSIM and EFPA are faster than LTL for the Atlas robot.
However, if we assume that $J$ is computed elsewhere and is available for re-use, its computation cost can be excluded from LTL operation count. Then our findings would concur with \cite{wensing2012reduced}, where LTL would be faster than EFPA for Atlas with 18 or 24 constraints, but still slower than the PV-OSIM. 
For Talos, LTL was not competitive with the lower order methods especially due to the expense of computing and factorizing a bigger JSIM.

The PV-OSIM-fast avoids computing and factorizing the dense inverse OSIM matrix explicitly using the matrix inversion lemma, and scales better than the PV-OSIM as the size of the OSIM matrix increases. It is the fastest algorithm for the considered floating-base robots and even nearly 2x faster than the LTL for the humanoid robots.

Though the PV-OSIM was computationally faster than the EFPA for all the considered robots, the EFPA has a lower order computational complexity of $O(n + md + m^2)$ compared to the $O(n + m^2d + m^2)$ of the PV-OSIM for computing the inverse OSIM $\Lambda^{-1}$.
This would make EFPA scale better than PV-OSIM for longer mechanisms with many constraints. 
To test this, we consider a long-stemmed mechanism ($n_{\mathrm{stem}}$ is the number of links in the stem). From both stem ends, $m_\mathrm{branches}$ chains of 7 links each branch out as shown in \cref{fig:OSIM_stalk}. Each branch's tip link is fixed with a 6D weld constraint. 

Figure~\ref{fig:OSIM_long_tree} shows the computational scaling of the ratio of PV-OSIM and EFPA operation counts w.r.t to $n_{\mathrm{stem}}$ for different values of $m_\mathrm{branches}$. 
EFPA was found to be always slower than PV-OSIM for up to 8 branches ($8 \times 6$ constraints propagated) irrespective of $n_{\mathrm{stem}}$. For 9 or more branches, the EFPA eventually becomes more efficient than PV-OSIM at a cross-over point stem length $n_{\mathrm{stem}}$. The value of the crossover point depends on $m_\mathrm{brances}$ as well as the branches' link length for the considered mechanism. More branches would reduce the cross-over point as EFPA can more efficiently propagate large number of  constraints through the stem links. Shorter branch length can also reduce the cross-over point because the constraint propagation through the stem links (where EFPA is more efficient than PV-OSIM) will form a fraction of the computations.  For a $m_\mathrm{branches} = 10$, the cross-over $n_\mathrm{stem} = 54$ for branch length of 7, which is a very large mechanism with $54 + 10 \times 7 \times 2 = 194$ links. For an extreme branch length of only 1 link, the cross-over $n_\mathrm{stem}$ can be as small as 7.  
%For EFPA to be more efficient than PV-OSIM due to its lower order complexity, many constraints need to be propagated through a large fraction of a mechanism's links. Consider the extreme case of each branch being 1 link long with a 6D weld constraint on this link and a mechanism has 10 branches on each side, here the crossover point is just 7 links in the stem.
Based on these findings, we conclude that the PV-OSIM requires fewer operations for most realistic robot mechanisms unless one is considering a heavily constrained mechanism with most constraints propagated through a large fraction of the joints.   

% \textcolor{red}{Discuss the connection with handles.}

%For all these different tasks compare the PV method with the textbook method for constrained dynamics. Also compare with either Pinocchio or the new paper by Justin Carpentier. For just the dynamics.And then also compare the derivative computation times for these methods.

\subsection{Benchmarking constrained dynamics solvers}

We compared the PV solver, PV-e solver and the PV-s solver with the state-of-the-art sparsity exploiting LTL solver of Featherstone \cite{featherstone2005efficient}, \cite{featherstone2010exploiting}. The LTL-OSIM \cite{featherstone2010exploiting} solver is a popular algorithm implemented in the high-performance simulator software \pkg{Pinocchio} \cite{carpentier2019pinocchio}. The LTL solver is also used in \pkg{MuJoCo} \cite{todorov2012mujoco} which uses a joint-space version of the soft-Gauss principle. To make a fair comparison with the LTL solvers, we implemented them ourselves and \cref{table1} reports the computation time taken by the different algorithms. The type and the number of constraints imposed are reported next to the robot name in parentheses. 

The computation times for the nominal \pkg{C++} and \pkg{C} execution of \pkg{Pinocchio} (Pin) and \pkg{MuJoCo}  (Mu) respectively cannot be considered a fair comparison because they do not use code-generation (which prunes unnecessary computations) and may compute additional quantities that are not required for constrained dynamics. We still report their computation timings for reference and indicative purpose of the speed-ups these software may achieve by exploiting code-generation.

\begin{table}[t]
  \vspace{0.2cm}
  \caption{Benchmarking computational performance of PV solver with other constrained dynamic solvers in \pkg{MuJoCo} and \pkg{Pinocchio}. All times are in microseconds. \label{table1}}
  \centering
  \begin{tabular}{c | c c c c|c c c}
    \hline
    Robot        & PV            & PV-e           & LTL      & Pin$^*$& PV-s         & LTL-s& Mu$^*$   \\
    \hline
    Iiwa (0D)    & \textbf{0.55} & \textbf{0.55}  & 0.63     & 2.15  & \textbf{0.55} & 0.63 & 3.11 \\
    Iiwa (H$_3$) & 0.75          & \textbf{0.61}  & 0.83     & 2.73  & \textbf{0.61} & 0.80 & 4.45 \\
    Iiwa (H)     & \textbf{1.01} & 1.09           & 1.08     & 3.53  & \textbf{0.63} & 0.89 & 4.88  \\
    Go1 (0D)     & \textbf{1.65} & \textbf{1.67}  & 1.74     & 4.68  & \textbf{1.64} & 1.74 & 7.10 \\
    Go1 (F$_3$)  & 1.88          & \textbf{1.81}  & 1.96     & 5.61  & \textbf{1.70} & 1.84 & 11.2 \\
    Go1 (2F$_3$) & 2.10          & \textbf{1.98}  & 2.20     & 6.40  & \textbf{1.76}& 1.98 &  12.0\\
    Go1 (3F$_3$) & 2.32          & \textbf{2.14}  & 2.48     & 7.33  & \textbf{1.82} & 2.16 & 12.8 \\
    Go1 (4F$_3$) & 2.53          & \textbf{2.33}  & 2.85     & 8.20  & \textbf{1.90} & 2.33 &  13.5\\
    Atlas (0D)   & \textbf{3.44} & \textbf{3.47}  & 4.64     & 12.3  & \textbf{3.47} & 4.64 & 15.9  \\
    Atlas (F)    & 4.59          & \textbf{3.94}  & 5.88     & 15.4  & \textbf{3.61} & 5.58 & 31.5 \\
    Atlas (2F)   & 6.09          & \textbf{4.40}  & 7.52     & 18.5  & \textbf{3.73} & 6.61 & 34.2 \\
    Atlas (2F+H) & 7.37          & \textbf{5.03}  & 8.69     & 22.3  & \textbf{3.76} & 6.93 & 36.5 \\
    Atlas(2F+2H) & 8.27          & \textbf{5.52}  & 11.8     & 26.5  & \textbf{3.82} & 7.77 & 38.8\\
    Talos (0D)   & \textbf{4.92} & \textbf{4.97}  & 8.14     & 17.1  & \textbf{4.96} & 8.28 & 23.6 \\ 
    Talos (F)    & 5.63          & \textbf{5.48}  & 9.25     & 21.1  & \textbf{4.96} & 8.65 & 51.3    \\
    Talos (2F)   & 6.72          & \textbf{6.45}  &  10.9    & 25.2  & \textbf{4.99} & 9.21 & 54.3    \\
    Talos (2F+H) & 8.40          & \textbf{7.06}  &   13.4   & 30.0  & \textbf{5.08} & 10.6 & 57.0   \\
    Talos(2F+2H)& 10.13          & \textbf{7.40}  &   15.4   & 34.7  & \textbf{5.11} & 11.9 &  59.4   \\
    %Atlamal (6D) \\
    %Atlamal (12D) \\

    \hline
  \end{tabular}
  $^*$ Pin and Mu are nominal execution of \pkg{Pinocchio} and \pkg{MuJoCo} without code-generation and hence cannot be considered fair comparison. 
\end{table}

\subsubsection{Hard motion constraints}

The PV-solver was as fast or faster than the sparsity-exploiting LTL methods for all the considered robots. The difference, while negligible for the 7 d.o.f Iiwa robot, widens for larger robots and more constraints due to its lower order complexity. Our PV-e solver scales even better than the PV solver, due to its lower order complexity of $O(n + m)$. For larger robots like Atlas or Talos with a high number of constraints, PV-e offers nearly a 50\% and 30\% reduction in computation compared to LTL and the PV-solver respectively.

%For a Kuka Iiwa robot with a 6D constraint on the end-effector, which is a typical scenario during operational-space control, PV solver and PV-s solver take 0.76 microseconds and 0.53 microseconds respectively while MuJoCo and Pinocchio take 3.79 microseconds and 3.21 microseconds respectively. 

%\textcolor{red}{ While it is difficult to claim that the proposed PV solver algorithm is faster than Pinocchio's algorithm since the code-generation environments are different.}

\subsubsection{Soft constraints}

The last three columns of the \cref{table1} present the computation times of our PV-s solver (see \cref{sec:soft_gauss}), our implementation of the \pkg{MuJoCo}'s soft Gauss principle using LTL and the nominal \pkg{C} execution in \pkg{MuJoCo} itself.
In  \pkg{MuJoCo}, we imposed 6D {\tt weld}-type equality constraints for F or H and 3D {\tt connect}-type equality constraints for F$_3$ and H$_3$ respectively. We deactivated all other constraints and frictional contacts (turned on by default in \pkg{MuJoCo}) to ensure that it solves the same equality constrained dynamics problems.
The PV-s implementation is significantly faster than all the other algorithms. It is nearly twice as fast as LTL-s and nearly thrice as fast as LTL (which arguably solves harder problem with hard motion constraints). It is unlikely that any constrained dynamics algorithm, that we know of, can compete with PV-s since its computation cost is nearly the same as that of the ABA algorithm (unconstrained forward dynamics algorithm with $O(n)$ complexity).

\begin{figure}

  \begin{subfigure}[h]{0.80\linewidth}
    \centering
    \resizebox{\linewidth}{!}{\input{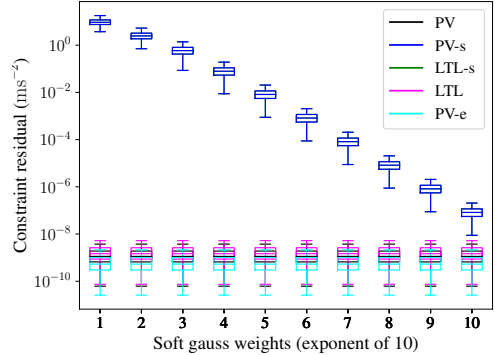}}
    \caption{Constraint residuals for different soft Gauss weights.}
    \label{fig:con_residuals}
    \end{subfigure}

  \begin{subfigure}[h]{0.80\linewidth}
  \centering
  \resizebox{\linewidth}{!}{\input{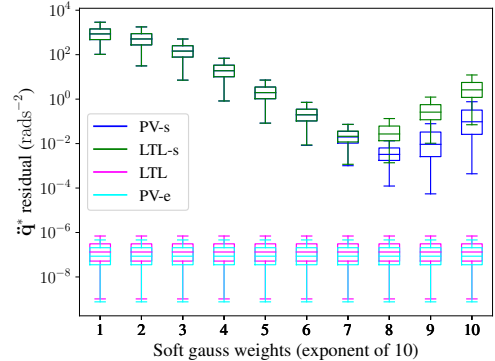}}
  \caption{Residuals of $\mathbf{\ddot{q}}^*$ w.r.t PV}
  \label{fig:qdd_residuals}
  \end{subfigure}

  \caption{Benchmarking the numerical accuracy of soft Gauss solver for different weights.}
  \label{fig:soft_gauss_accuracy}
\end{figure}

\subsubsection{Accuracy of the proposed solvers}

We benchmarked the accuracy of the soft Gauss principle for different value of weights in \cref{fig:soft_gauss_accuracy}. 
We present the whisker plots of $\ell_2$ norm of the constraint residuals in \cref{fig:con_residuals} and the $\ell_2$ norm of the difference in $\mathbf{\ddot{q}}^*$ computed by the PV solver (reference algorithm because it considers hard motion constraints) in \cref{fig:qdd_residuals} for the Talos robot with 2H+2F constraint (both its feet and hands are fixed with a full 6D constraint) at 1000 different randomly sampled joint configurations. PV, PV-e and LTL that solve for hard equality constraints satisfy the constraint to high level of accuracy, with PV-e appearing to be numerically slightly stabler than the other two. Both the soft Gauss solvers, PV-s and LTL-s, have a significantly higher value of constraint residual, though the residual keeps reducing as the penalty weights are increased. Both PV-s and LTL-s satisfy the constraints equally well. However, for weights higher than a certain point ($\thicksim 10^8$), the optimal joint accelerations computed by the soft Gauss solvers and the hard Gauss solvers begin to diverge due to numerical issues, where the high penalty weights begin to affect the joint acceleration solution in the nullspace of the constraints. Between the two soft Gauss solvers, PV-s appears to be more numerically stable than LTL-s. % perhaps inheriting the well-known improved numerical stability \cite{featherstone2004empirical} of a similar propagation-based method ABA w.r.t to the LTL algorithm for unconstrained dynamics. 

%MuJoCo is a feature-rich solver that supports unilateral (inequality) constraints and frictional contacts
%Please note that the soft version of the PV solver (PV-s) and MuJoCo (Mu) are solving the soft Gauss' principle which is computationally an easier problem than solving for hard constraints which the original PV solver (PV) and Pinocchio (Pin) do. 

%Based on these results, we observe that code-generation is a highly effective tool to speed up dynamics solvers irrespective of the algorithm used. The compilation of the C-code generated by CasADi took \textcolor{red}{under 2 seconds} even for the 18 d.o.f quadruped with 12 constraints. 

\subsection{Computational scaling} \label{subsec:scaling}

    \begin{figure}
      \begin{subfigure}{.95\linewidth}
      \centering
      \resizebox{\linewidth}{!}{\input{graphics/chain_scaling_log.pgf}}
      \caption{Computational scaling for chains with fixed-base and 6D constrained end effector.}
      \label{fig:scaling_logplot}      %     \label{fig:scaling_plots}
    \end{subfigure}\hspace{0pt}

    \begin{subfigure}{.95\linewidth}
      \centering
      \includegraphics[width=0.5\textwidth]{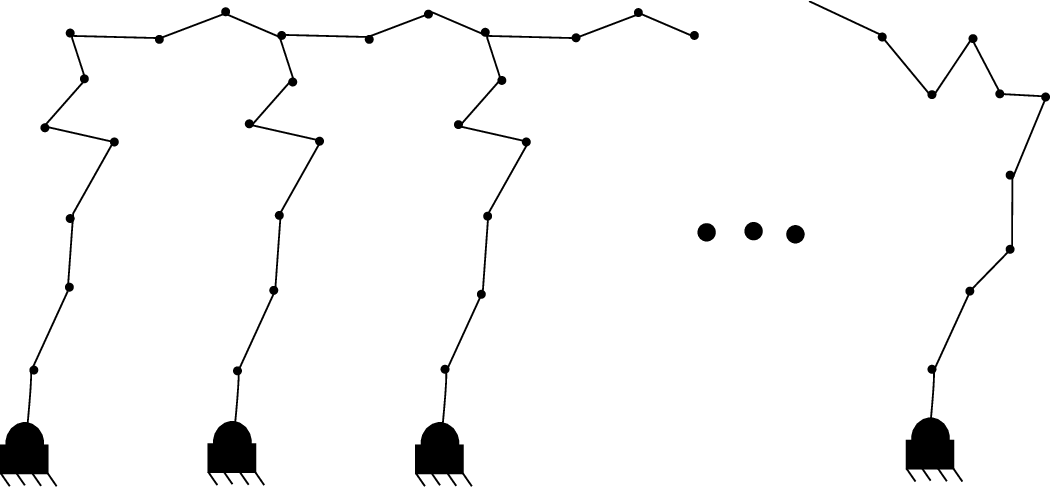}
      \caption{The ladder mechanism where $m \thicksim O(n)$.}
      \label{fig:ladder_mechanism}
    \end{subfigure}\hspace{0pt}

      \begin{subfigure}{.95\linewidth}
        \centering
        \resizebox{\linewidth}{!}{\input{graphics/ladder_scaling.pgf}}
        \caption{Benchmarking computational scaling for the ladder mechanism.}
        \label{fig:scaling_trees}
      \end{subfigure}\hspace{0pt}
      \caption{Computational scaling of the different algorithms.}
      \label{fig:computational scaling} 
    \end{figure}

We empirically tested the computational scaling of the different constrained dynamics algorithms and present the results in \cref{fig:computational scaling}. In \cref{fig:scaling_logplot}, we show computational times of the different algorithms for kinematic chains ranging from 6 to 100 revolute joints. The end-effectors are fixed with full 6D constraints. As expected, the $O(n)$ complexity PV, PV-e and PV-s solvers scale linearly and more gracefully than the higher-order LTL and LTL-s algorithms used in \pkg{Pinocchio} and \pkg{MuJoCo} respectively. Beyond a certain number of links, the generated C-code for LTL and LTL-s became too large for effective compiler optimization and they became slower than even the nominal \pkg{C++} execution in Pinocchio. %The figures show an unexpected jump in computation timings for MuJoCo at around 60 links, which was found to be repeatable over different runs. We suspect it may be caused by MuJoCo's program becoming too large to use cache memory efficiently. 

Then we compared the different algorithms on a highly constrained ladder-shaped mechanism (see \cref{fig:ladder_mechanism}) with $m \thicksim O(n)$, with each rung consisting of 7 links. The segment connecting two ends of a rung on one side has 3 links and the other ends of the rung are constrained to be fixed with full 6D constraints. The computational timings of different algorithms as more rungs (and constraints) are added to the mechanism are presented in \cref{fig:scaling_trees}. The PV solver with its cubic complexity in the number of constraints also begins to scale badly like the LTL and LTL-s solvers, while the $O(m + n)$ solvers PV-e and PV-s scale linearly.

\subsection{Discussion and limitations}

\label{sec:discussions_limitations}

% Parallels with OCP solvers, 
% ABA is condensing \cite{bock1983recent}
% PV solver is partial condensing
% PV-e is equality constrained LQR.
\subsubsection*{Parallel algorithms} Our comparison was limited to implementations on a single core. However, the divide-and-conquer algorithms \cite{featherstone1999dividea,featherstone1999divideb,yamane2009comparative,bhalerao2012efficient} may be computationally faster, especially for bigger mechanisms, when multiple cores are utilized. On a single core however, they are unlikely to be faster for typical robots since they are known to be several times more expensive than ABA \cite{featherstone1999divideb}. However, due to a lack of open source implementation and due to the complexity of their implementation, we leave this comparison for future work.

Among these divide-and-conquer methods the PV solver appears to be most closely related to the DCAp algorithm \cite{featherstone1999divideb}, which has outward acceleration propagation and inward force propagation similarly to the PV solver and the ABA is shown to be a special case of DCAp. It appears to be possible to provide an alternative derivation of the PV solver from the DCAp algorithm by placing a handle on the floating-base and the constrained links. The handles on the constrained links would be in the constraint space instead of the spatial handle explicitly considered in \cite{featherstone1999divideb}. Then, using the two-handle equation in \cite[sec. 4.1]{featherstone1999divideb}, for a specific order of assembly from the leaf nodes to the root, it is possible to show that \cite[eq. 29a, 29g, 29b, 29h, 29d]{featherstone1999divideb} correspond to \cref{eq:tree_H_recursion}, \cref{eq:tree_f_recursion}, \cref{eq:tree_K_recursion},  \cref{eq:tree_l_recursion} and \cref{eq:tree_L_recursion} respectively. However, such an assembly ordering is not the recommended ordering in divide-and-conquer algorithms as it does not assemble two trees of similar sizes which is necessary for obtaining a reduced order complexity in the divide-and-conquer methods.

Though there is no direct analogue for the PV-early algorithm in DCAp, a simpler form of early elimination can also be performed in DCAp when the $L^A$ matrix reaches full rank by eliminating the constraint forces by taking Schur complement. Due the divide-and-conquer methods being among the most complex rigid-body dynamics algorithms in literature, deriving the PV solver this way may not be of interest to readers. However, this connection opens up interesting possibilities for parallelizing the algorithm, which we leave for future work.

\subsubsection*{Closed-loop solvers} The PV solver is closely related to the algorithms in \cite{otter1987algorithm} and \cite{bae1987recursive}. In the PV solver's backward recursion, the \cref{eq:constraint_force_prop}, \cref{eq:acc_setpoint_prop} and \cref{eq:OSM_Recursion} correspond to \cite[eq. 16c, eq. 18a and eq. 18b]{otter1987algorithm} respectively and \cite[eq. 41c, eq.51b, eq.51a]{bae1987recursive} respectively. Application-wise, the main difference between PV-solver and \cite{otter1987algorithm,bae1987recursive} is that we consider known acceleration constraints (which includes all the loop closure constraints with the ground as a special case), while both \cite{otter1987algorithm,bae1987recursive} tackle the harder problem of internal kinematic loop constraints. We also explicitly consider  floating-base systems which was not considered in \cite{otter1987algorithm}, while \cite{bae1987recursive} does consider floating-base systems in one of their examples though not in the main derivation. Both \cite{otter1987algorithm} and \cite{bae1987recursive} can be straight-forwardly adapted to solve the constrained dynamics problems considered by the PV solver. This connection between the PV solver, \cite{otter1987algorithm} and \cite{bae1987recursive} appears to not have been made in existing literature. Despite not being a fundamentally new algorithm, the expository PV solver derivation in \cref{sec:derivation_chain} and \cref{sec:derivation_tree} is of value to the readers because it utilizes a different LQR perspective that permitted a mechanistic derivation of the algorithms, that would make the material accessible to researchers with control and optimization background. In contrast, \cite{otter1987algorithm} required significant physical insight to come up with an efficient propagation of Newton-Euler solutions similarly to the ABA algorithm \cite{featherstone1983calculation}. However, \cite{otter1987algorithm} approach may be more accessible to researchers with a background in mechanics and without prior experience in optimal control or optimization.

\subsubsection*{$O(n + m)$ solvers} Our expository derivation also allowed us to easily derive two different and original (to the best of our knowledge) $O(n + m)$ solvers, using the soft Gauss principle adopted by \pkg{MuJoCo} and early elimination of dual variables. A form of early elimination is also proposed in \cite{otter1987algorithm,bae1987recursive}, where they eliminate the dual variables of a loop after passing over all the links in that loop. For certain robot architectures where the loops are not heavily interconnected (the same link being part of multiple loops), their early elimination procedure can also lead to $O(m+n)$ performance.  Our early elimination is fundamentally different as it reduces the dimensionality of the propagated constraints at every joint. 

A relatively more recent $O(n + m)$ complexity solver for kinematic loops \cite{anderson2003improvedorder} uses the same ideas as \cite{otter1987algorithm} by introducing zero-mass phantom link for loop-cutting and early elimination at the loop level. However, unlike \cite{otter1987algorithm} and the PV solver, \cite{anderson2003improvedorder}  proposes a Lagrange multiplier free algorithm based on Kane's formulation of constrained dynamics \cite{kane1985dynamics}. The algorithm in \cite{anderson2003improvedorder} is fairly complex, does not have an open-source implementation and does not appear to have been benchmarked with the PV solver, \cite{otter1987algorithm} or \cite{bae1987recursive}. It is not obvious how to efficiently adapt it to the kinematic-tree structures considered by the PV-solver. Despite  \cite{anderson2003improvedorder} being a challenging algorithm to understand and implement, the Lagrange multiplier-free approach is interesting and may be computationally beneficial, especially for mechanisms with kinematic loops, and will be investigated in the future.  %Our early elimination algorithm in its current version is not suitable for kinematic loops, but using the local coordinate and phantom-body ideas introduced in \cite{anderson2003improvedorder}, it can be efficiently extended to kinematic loops as well, which we leave for future work.

The SVD currently proposed for PV-early is admittedly an expensive algorithm for multi d.o.f joints, when we cannot exploit the efficient rank-1 update formulae presented in \cref{sec:early_multiplier}, unless the multi d.o.f joints are modelled as several equivalent fictitious single d.o.f joints in a chain. However, this workaround is non-ideal as it introduces issues like representation singularity and non-physical meaning of velocities of these fictitious joints. It may be worthwhile to explore replacing the SVD with the more efficient rank-revealing QR decomposition \cite{golub2013matrix} in the future, which provides the desired orthogonal bases similarly to the SVD. 

\subsubsection*{OSIM and computational benchmarking} That the backward recursion in PV solver, \cite{bae1987recursive} and \cite{otter1987algorithm} provides an efficient algorithm to compute the OSIM is a new connection made in this paper that we could not find in literature. We are also not aware of existing work that computationally benchmarked the PV-solver or the \cite{otter1987algorithm}, \cite{bae1987recursive} algorithms with the currently popular sparsity-exploiting methods of Featherstone for the constrained dynamics problems considered in this paper. Our findings indicate that for larger robots like the humanoid robots the sparsity-exploiting methods are not competitive with the PV solver, which has implications for the existing simulators and as well as for biomechanical applications where the degrees of freedom are typically over 100.

Our benchmarking methodology included code-generating and compiling robot-specific C code, which while contributing to the speeds we observe, is also a limitation as we need to know all the possible contact situations that may arise. Nominal \pkg{C++} implementations such as in \pkg{Pinocchio} can deal with these scenarios more effectively as they do not require re-compilation at runtime. However, in many applications e.g. humanoid walking, all the possible contact scenarios can be compiled in advance and loaded depending on the contact scenario using look-up tables. In any case, the speed-up we observed due to code-generation is high enough that it is interesting for simulators to explore a hybrid method combining the strengths of both code-generation and nominal \pkg{C++} execution for different parts of the algorithm.

Finally, we refer interested readers to several extensions of the unconstrained LQR algorithm to equality-constrained problems \cite{park2008lq,giftthaler2017projection,laine2019efficient,vanroye2023generalization} in a control setting. Out of these methods \cite{park2008lq} is analogous to the original PV solver and \cite{laine2019efficient}'s method is most similar to our PV-early solver, where they also used SVD. %However, our method can make stronger assumptions due to the positive definiteness of the articulated body inertia, whcih    While there are several constrained algorithms proposed in the control community \cite{laine2019efficient},

%It remains to be seen if this advantage of PV-s can extend to more complex constraints, like unilateral constraints and frictional contacts, supported by a feature-rich simulator like \pkg{MuJoCo}. Efficient extension of PV-s to these complex constraints is in itself a research problem and is therefore outside the scope of this paper.

% The code-generated simulation is also nearly an order of magnitude faster for the nominal execution of the MuJoCo.

%\textcolor{red}{Both sparsity exploiting and linear complexity}

%\textcolor{red}{Talk about potential parallelizability or leave it to future work.}

\section{CONCLUSIONS AND FUTURE WORK}

\label{sec:Conclusions}

\subsection{Conclusions}

%dynamic programming treatment of constrained dynamics by connecting to LQR. Extending the connections to LQR found by Rodriguez, but much simpler derivation using dynamic programming. People in MPC using robot dynamics (especially constrained dynamics) that use it as a black-box, this reveals the internal details of the algorithms.

%Updated treatment of the connections made by Rodrigues for motion constraints.

We provided a self-contained derivation of several advanced constrained dynamics solvers from the first principles by connecting it to the LQR problem. 
Our derivation, building upon Vereshchagin's approach, is much simpler than the better known SOA framework of Rodriguez \cite{rodriguez1987kalman} that uses this LQR connection. 
Our expository derivation extended the original PV solver to floating-base kinematic trees, which resulted in an algorithm closely related to \cite{bae1987recursive} and \cite{otter1987algorithm}, but is derived using a different LQR perspective.
This paper makes constrained dynamics accessible to researchers in optimization and control as well as roboticists, with knowledge of control, that currently treat robot dynamics as a black-box and are therefore unable to debug or adapt existing dynamics software to their applications. The LQR connection can foster transfer of software and ideas between fields in the future. For example, recent research from  data-driven LQR control may transfer to robust control of robots with uncertain dynamics. 
The optimization perspective in our derivation is valuable as accounting for uncertainty in parameters is performed naturally in an optimization framework \cite{ben2009robust}, \cite{shapiro2021lectures}. 

The equality we showed between LQR's dual Hessian and the inverse OSIM provided an efficient state-of-the-art OSIM algorithm, which we further significantly accelerated for specific, but common, robot structures that have branching at the base. The LQR-based approach allowed straightforward derivation for the PV-s and PV-early algorithms, resulting in two original algorithms with $O(n + m)$ complexity. Our numerical experiments suggest that the PV solver is computationally superior to currently popular higher-order sparse factorization algorithms by Featherstone for larger robots like the humanoid robot Atlas, for which the LTL needs up to 2x more computations than the PV-solver. This PV-solver speed-up can be arbitrarily higher for longer mechanisms, typical in biomechanical applications, due to the inherent complexity difference. Finally, our work recognizes the historical contribution of Popov and Vereshchagin who proposed the first $O(n)$ \textit{constrained} dynamics solver, which remarkably remains the state-of-the-art nearly fifty years after its invention and yet remains largely unknown in the robotics community.

\subsection{Future work}

There are multiple exciting directions for future work, apart from the applications in robot control and trajectory optimization. 
The algorithms presented here are limited to equality constraints, and it is a natural research direction to extend the algorithms to include internal kinematic loops, frictional contacts and unilateral contact constraints. We will also explore proximal point iterations \cite{carpentier2021proximal} for applying the solver to problems with ill-conditioned and nearly redundant constraints.
Analytical gradients, which are found to be faster than automatic differentiation, can also be developed for the PV solver for optimal control and reinforcement learning applications. 
In particular, transfer of new research results from data-driven LQR to robot control is an exciting future research direction.

\section*{Acknowledgement}
The authors thank Prof. Jan Swevers, Bastiaan Vandewal and Alejandro Astudillo Vigoya for their valuable feedback on previous versions of the manuscript. The authors also thank the anonymous reviewers for their valuable comments and suggestions. We especially thank the anonymous reviewer 1 for the extensive review and for pointing us to important literature that we were not aware of (e.g. Brandl et al.'s paper).

\bibliographystyle{IEEEtran}
\bibliography{references}

\vspace*{1\baselineskip}

\begin{wrapfigure}{l}{25mm} 
  \includegraphics[width=1in,height=1.25in,clip,keepaspectratio]{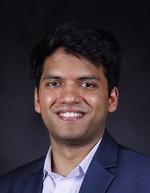}
\end{wrapfigure}\par
{\small \textbf{Ajay Sathya} obtained his Bachelors degree and Masters degree from NITK Surathkal, India and KU Leuven, Belgium respectively. He is currently pursuing a PhD degree at Mechanical Engineering department of KU Leuven, Belgium.  \par }

\vspace*{4\baselineskip}

\begin{wrapfigure}{l}{25mm} 
  \includegraphics[width=1in,height=1.25in,clip,keepaspectratio]{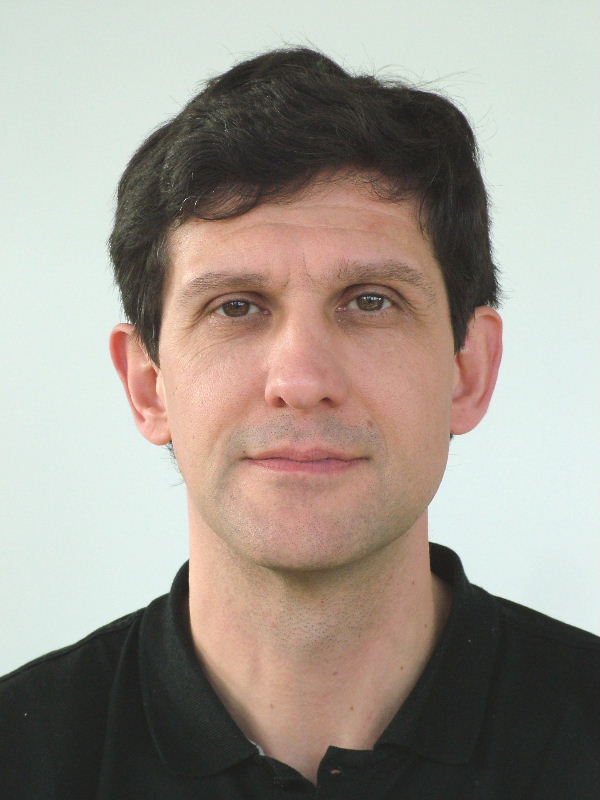}
\end{wrapfigure}\par
{\small \textbf{Dr. Bruyninckx}  
(\href{https://u0011821.pages.mech.kuleuven.be/}{Personal webpage}) obtained the Masters degrees in Mathematics (Licentiate, 1984),
Computer Science (Burgerlijk Ingenieur, 1987) and Mechatronics (1988), all from the KU Leuven, Belgium. In 1995 he obtained
his Doctoral Degree in Engineering from the same university.
He is full-time Full Professor at the KU Leuven, and partime at the Eindhoven University of Technology. The research focus
in both places is on the composability of the most advanced, knowledge driven algorithms for the dynamics of motion control
of complex robotics applications, with distributed sensor processing and resource monitoring. The complementary objectives
are to realise such systems with the least amount of resource costs, with ``good enough'' quality, and with full
``explainability''.\par}

\vspace*{1\baselineskip}

\begin{wrapfigure}{l}{25mm} 
  \includegraphics[width=1in,height=1.25in,clip,keepaspectratio]{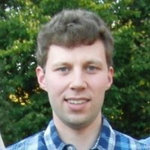}
\end{wrapfigure}\par
{\small \textbf{Wilm Decr\'e} (Member, IEEE) received the B.S.,
M.S., and the Ph.D. degrees in Mechanical Engineering
from KU Leuven, Belgium in 2004, 2006,
and 2011, respectively. He is a research manager at
the Department of Mechanical Engineering of KU
Leuven, Belgium.
His research interests include sensor- and
optimization-based control of robot systems, numerical
optimization algorithms and applications,
learning and optimal control and estimation, and
real-time and embedded software design.
 \par}
 
\vspace*{\baselineskip}

\begin{wrapfigure}{l}{25mm} 
  \includegraphics[width=1in,height=1.25in,clip,keepaspectratio]{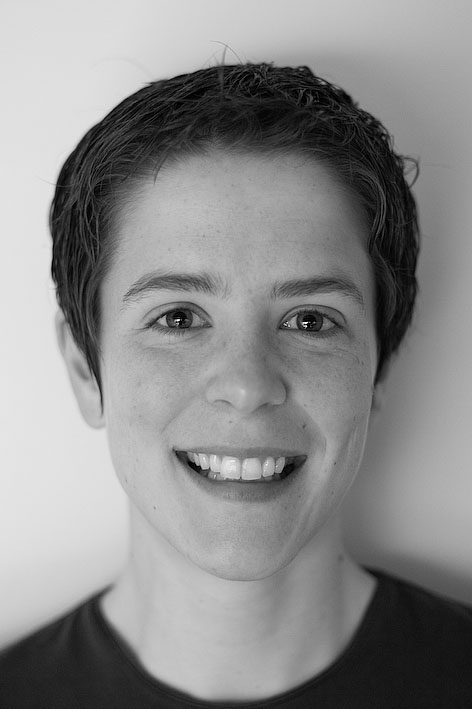}
\end{wrapfigure}\par
{\small 
After an academic career at the  KU Leuven Department of Mechanical Engineering, \textbf{Goele Pipeleers} moved to Materialise N.V., where she currently focusses on innovations in additive manufacturing.\par}

\balance
\end{document}